\documentclass[sigconf]{acmart}
\usepackage{multirow}
\AtBeginDocument{%
  }

\setcopyright{acmlicensed}
\copyrightyear{2018}
\copyrightyear{2025}
\acmYear{2025}
\setcctype{by}
\acmConference[SIGSPATIAL '25]{The 33rd ACM International Conference on Advances in Geographic Information Systems}{November 3--6, 2025}{Minneapolis, MN, USA}
\acmBooktitle{The 33rd ACM International Conference on Advances in Geographic Information Systems (SIGSPATIAL '25), November 3--6, 2025, Minneapolis, MN, USA}
\acmDOI{10.1145/3748636.3762776}
\acmISBN{979-8-4007-2086-4/2025/11}

\usepackage{times}
\usepackage{soul}
\usepackage{url}
\usepackage[utf8]{inputenc}
\usepackage{caption}
\usepackage{graphicx}
\usepackage{booktabs}
\usepackage{algorithm}
\usepackage{subfig}   % Or the `subcaption` package
\usepackage{algorithmic}
\usepackage{multirow}
\usepackage[switch]{lineno}
\usepackage{color}
\usepackage{tabularx}
\usepackage{xcolor}
\usepackage{enumitem}
\newcommand{\DEL}[1]{\iffalse #1 \fi}

\newtheorem{definition}{Definition}
\begin{document}
%\title{EAMTF: Event Aware Multimodal Traffic Forecasting}
%\title{LLM-GNN Fusion for Event-Aware Spatiotemporal Traffic Forecasting}
\title{FUSE-Traffic: Fusion of Unstructured and Structured Data for Event-aware Traffic Forecasting}

%%
%% The "author" command and its associated commands are used to define
%% the authors and their affiliations.
%% Of note is the shared affiliation of the first two authors, and the
%% "authornote" and "authornotemark" commands
%% used to denote shared contribution to the research.
\author{Chenyang Yu, Xinpeng Xie, Yan Huang, and Chenxi Qiu}
\email{{chenyangyu, xinpengxie}@my.unt.edu,{yan.huang, chenxi.qiu}@unt.edu}
\affiliation{%
  \institution{Department of Computer Science and Engineering, University of North Texas}
  %\streetaddress{P.O. Box 1212}
  \city{Denton}
  \state{Texas}
  \country{USA}
  %\postcode{43017-6221}
}

%%
%% By default, the full list of authors will be used in the page
%% headers. Often, this list is too long, and will overlap
%% other information printed in the page headers. This command allows
%% the author to define a more concise list
%% of authors' names for this purpose.
\renewcommand{\shortauthors}{Chenyang Yu, Xinpeng Xie, Yan Huang, and Chenxi Qiu}

%%
%% The abstract is a short summary of the work to be presented in the
%% article.
\begin{abstract}
Accurate traffic forecasting is crucial for Intelligent Transportation Systems (ITS) but is significantly challenged by non-periodic external events that disrupt regular traffic patterns. While Graph Neural Networks (GNNs) excel at modeling periodic traffic, they often falter in predicting event-driven dynamics. Existing event-aware methods either rely on manually engineered features with limited generalization or depend on curated textual event datasets that are costly to maintain and incomplete. The advent of Large Language Models (LLMs) offers new avenues for understanding and integrating event information. However, directly applying LLMs for all spatio-temporal reasoning can be inefficient, and effectively leveraging their event understanding capabilities within structured forecasting workflows remains a challenge. This paper introduces FUSE-Traffic, a framework which synergizes the dynamic event querying and understanding prowess of LLMs with the spatio-temporal modeling capabilities of GNNs. FUSE-Traffic features an on-demand event information extraction module using LLM prompting and a cross-attention based multimodal fusion mechanism to integrate rich event semantics with traffic flow features. This design enables the model to dynamically perceive and adapt to event-triggered traffic pattern changes. Comprehensive experiments on the METR-LA and PEMS datasets demonstrate that FUSE-Traffic significantly outperforms state-of-the-art models, especially under high-impact event conditions, showcasing robust predictive accuracy and resilience where traffic patterns are most disrupted. Code available at \url{https://github.com/GeoAICenter/FUSE-Traffic_Sigspatial2025}
\end{abstract}

%%
%% The code below is generated by the tool at http://dl.acm.org/ccs.cfm.
%% Please copy and paste the code instead of the example below.
%%


%%
%% Keywords. The author(s) should pick words that accurately describe
%% the work being presented. Separate the keywords with commas.

%%
%% This command processes the author and affiliation and title
%% information and builds the first part of the formatted document.

\maketitle

\section{Introduction}
\begin{figure}
    \centering
    \includegraphics[width=1\linewidth]{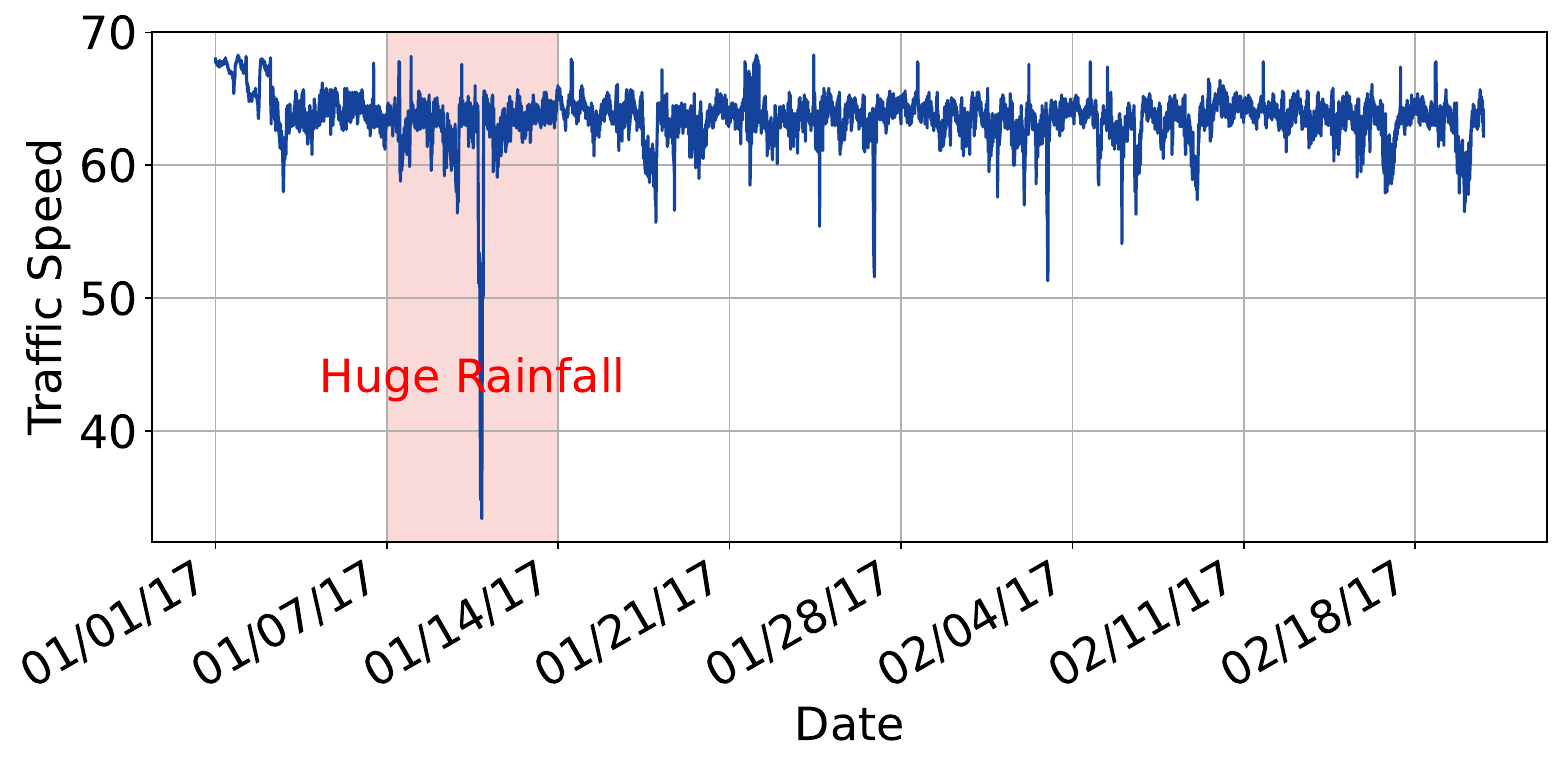}
    \caption{Event Impact on Traffic Status in PEMS-BAY}
    \label{fig:Event}
\end{figure}
Accurate traffic forecasting is a core technology for building Intelligent Transportation Systems (ITS), enabling better urban resource allocation and improved travel experiences.  With growing urbanization, traffic congestion has intensified, highlighting the need for reliable and responsive forecasting models. In recent years, deep learning, particularly Graph Neural Networks (GNNs), has emerged as the mainstream paradigm in traffic forecasting. GNNs can effectively capture complex spatial dependencies in road network topology and dynamic temporal evolution patterns in traffic flow data. Foundational models such as STGCN~\cite{Yu2018STGCN} and GraphWaveNet~\cite{Wu2019GraphWaveNet}, along with more recent developments including STWave~\cite{Fang2023STWave} and D2STGNN~\cite{shao2022decoupled}, have achieved impressive performance on standard traffic datasets. These approaches incorporate sophisticated graph convolutional structures and temporal modeling mechanisms, demonstrating particular effectiveness in capturing and forecasting traffic patterns characterized by periodic regularities.

However, real-world urban traffic systems are open, complex, and highly dynamic, where the operational states are not only driven by historical patterns but are also frequently and significantly affected by various non-periodic external events as shown in Figure \ref{fig:Event}. These events are wide-ranging, including traffic accidents, road construction, large-scale public activities (such as sports events, concerts ~\cite{Hong2024STORM}), extreme weather conditions, and even important social news. Such events often lead to sudden, irregular, and drastic fluctuations in traffic flow and speed, a challenge noted in early traffic studies~\cite{pan2012utilizing}. Traditional GNN models, relying solely on historical traffic data and fixed road network structures, often exhibit significant limitations in predicting and capturing these event-driven traffic dynamics. Therefore, effectively integrating external event information into prediction models to achieve event-aware traffic forecasting has become a key challenge for enhancing model robustness and practical application value. \looseness = -1

To address this challenge, researchers have explored various ways to incorporate event information. Early attempts primarily relied on manually engineered event features. For instance, some approaches introduced manually defined incident effect scores or constructed specific subgraphs for different event-induced traffic conditions~\cite{Xie2020DIGCNet, Luo2023M3AN}. While these methods somewhat enhance responsiveness to specific events, their core drawback lies in a heavy reliance on domain experts' prior knowledge, making generalization to diverse and complex unknown events difficult, and low-dimensional manual features often lead to the loss of rich semantic details.

\begin{figure*}[ht]
    \centering
    \includegraphics[width=0.87\textwidth]{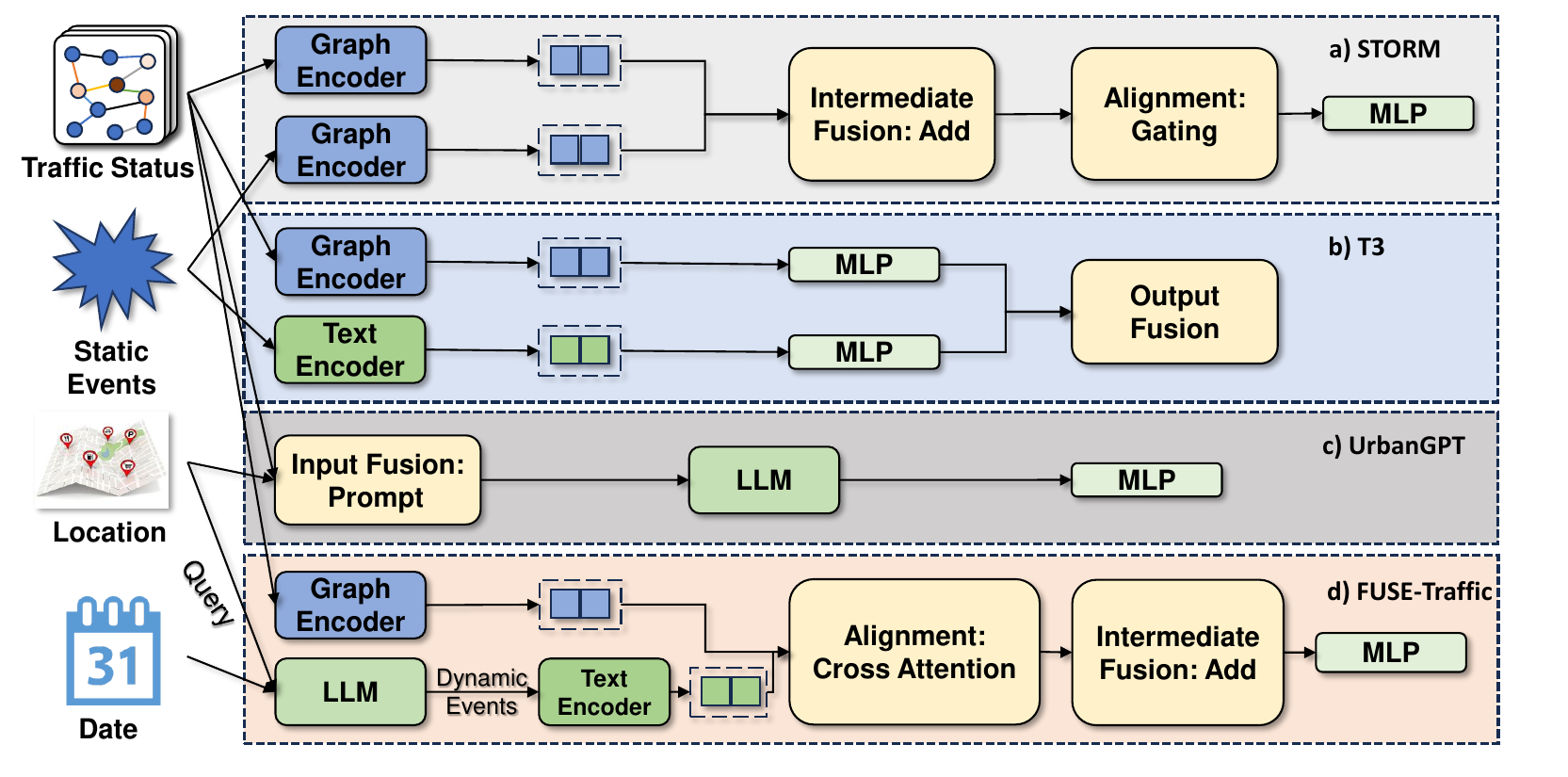}
    \caption{Overview of Four Representative Fusion Architecture. \emph{Top} to \emph{Bottom}:  \textbf{a)} STORM~\cite{Hong2024STORM} uses graph encoder to process event data and performs fusion \& Alignment between two modalities. \textbf{b)} T3~\cite{EventTraffic2024} performs fusion by concatenating graph and text representations at the output stage. \textbf{c)} UrbanGPT~\cite{Li2024UrbanGPT} incorporates spatiotemporal data and regional location information into the prompt, which is then fed into an LLM for prediction. \textbf{d)} FUSE-Traffic utilizes LLM for query events and jointly processes both graph and text representations via fusion and alignment}
    \label{fig:arch_types}
\end{figure*}

To overcome these limitations, subsequent research began to leverage textual descriptions of events. Figure~\ref{fig:arch_types} illustrates several multimodal architectural paradigms. One common strategy, exemplified by models like T3~\cite{EventTraffic2024} and visualized in Figure~\ref{fig:arch_types}(b), involves processing graph and text modalities separately and then fusing their representations, often through simple concatenation. A more sophisticated fusion approach, as adopted by works such as STORM~\cite{Hong2024STORM} and depicted in Figure~\ref{fig:arch_types}(a), employs mechanisms like cross-attention to enable deeper interaction between the modalities. While text-based models improve event sensitivity, they often depend on costly, curated event datasets that are difficult to maintain and generalize poorly to unseen scenarios. The construction and maintenance of such datasets are costly, and they can hardly cover all possible events affecting traffic.

In recent years, the rapid development of Large Language Models (LLMs), with their powerful natural language understanding, vast world knowledge reserves, and excellent in-context learning capabilities, has brought new light to event-aware traffic forecasting. This has led to explorations of directly using LLMs for spatio-temporal prediction. For example, as conceptualized in Figure~\ref{fig:arch_types}(c), UrbanGPT~\cite{Li2024UrbanGPT} encode various data types (e.g., traffic data, geographical information, time information) into natural language prompts, which are then fed into an LLM to directly output prediction results. This end-to-end approach can leverage the LLM's understanding of spatial temporal data. However, \emph{relying entirely on LLMs for all spatio-temporal reasoning can be computationally expensive, and LLMs may not be as adept as specialized GNNs in modeling fine-grained spatio-temporal structured dependencies}. More importantly, a critical gap remains in \emph{how to enable LLMs to dynamically query and accurately understand sudden public events based on real-time prediction needs, and to seamlessly and targetedly integrate this knowledge into structured traffic forecasting workflows}.

To address the aforementioned observations and challenges, in this paper we propose a novel framework, FUSE-Traffic. The core idea of FUSE-Traffic, whose architecture is overviewed in Figure~\ref{fig:arch_types}(d) and detailed in Figure~\ref{fig:arch}, is to leverage the dynamic event querying and understanding capabilities of LLMs and efficiently combine them with the spatio-temporal modeling capabilities of GNNs through a cross-attention mechanism. Specifically, we have designed a mechanism that enables an LLM to dynamically query and extract relevant external event texts based on the specific spatio-temporal context (e.g. local news/weather/crime) of the current traffic forecasting task. This semantically rich event information is then encoded and deeply fused with spatio-temporal features of traffic flow extracted by a graph encoder, utilizing a cross-attention mechanism. This design allows the model to dynamically perceive and adapt to traffic pattern changes triggered by different events, thereby improving prediction accuracy and robustness. The main contributions of this paper are as follows:

\begin{enumerate}
    \item We propose an innovative event-aware multimodal traffic forecasting framework, FUSE-Traffic, which effectively addresses the complex impacts of external events on traffic by synergizing LLMs (for dynamic event querying and understanding) with GNNs (for spatio-temporal traffic modeling).
    \item We design and implement an on-demand event information extraction module based on LLM prompting and a fusion and alignment mechanism, enabling fine-grained interaction and integration of event semantic information with spatio-temporal features of traffic. By dynamically integrating event semantics into the forecasting process, FUSE-Traffic adapts to non-periodic disruptions such as accidents, severe weather, or public gatherings, enabling more reliable and resilient traffic prediction in real-world scenarios.
    \item We conduct comprehensive experimental evaluations on three widely used public traffic datasets (METR-LA, PEMS-BAY and PEMS03). The results demonstrate that FUSE-Traffic significantly outperforms various state-of-the-art baseline models across multiple prediction horizons and evaluation metrics, especially under high-impact event conditions, showcasing robust predictive accuracy and resilience where traffic patterns are most disrupted.
\end{enumerate}

\section{Methodology}

\begin{figure*}
    \centering
    \includegraphics[width=0.99\linewidth]{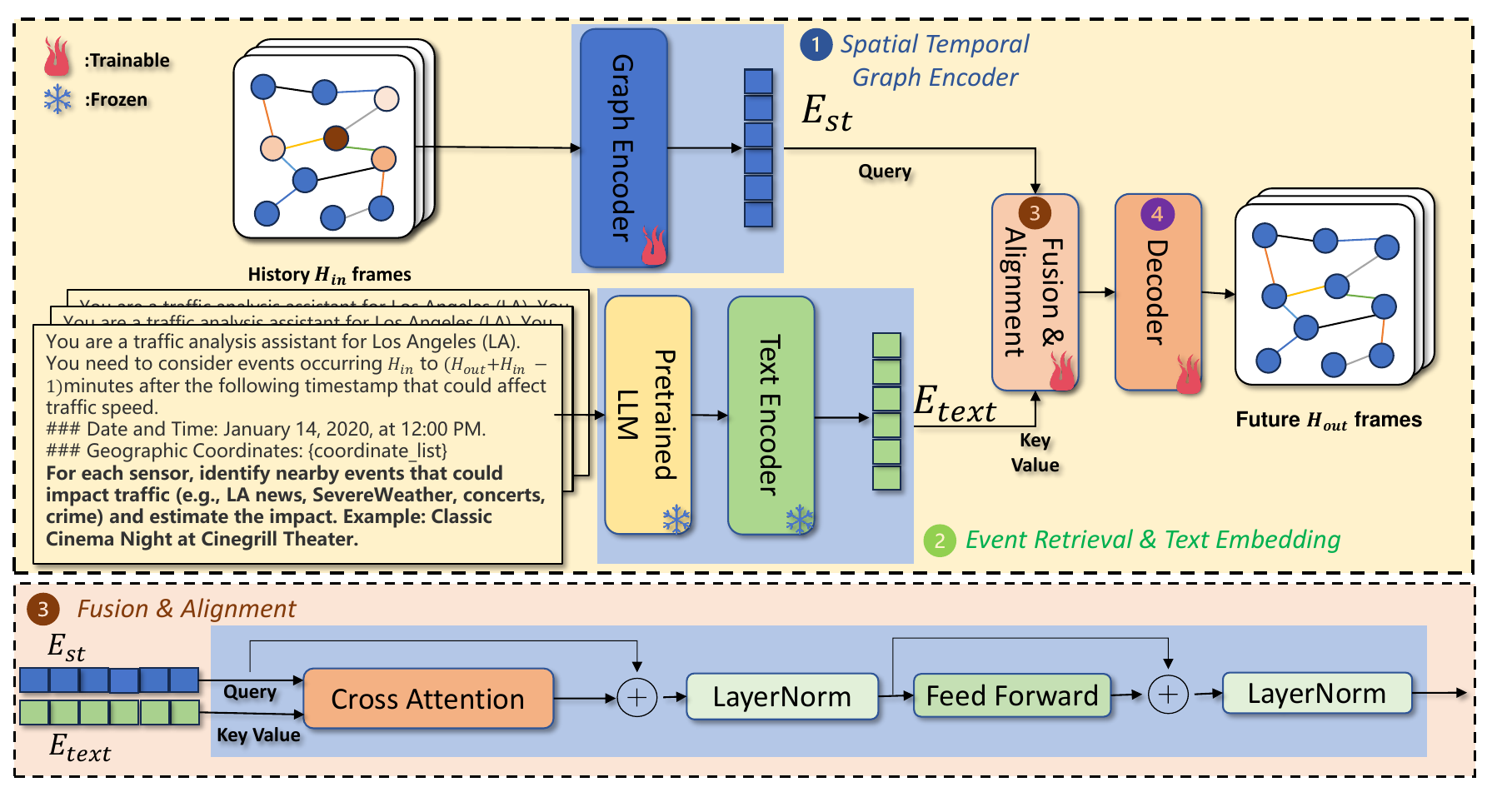}
    \caption{FUSE-Traffic Framework}
    \label{fig:arch}
\end{figure*}
In this section, we present our method, \emph{\underline{F}usion of \underline{U}nstructured and \underline{S}tructured data for \underline{E}vent-aware \underline{Traffic} forecasting framework (FUSE-Traffic)}. As shown in Figure \ref{fig:arch}, the FUSE-Traffic framework is composed of four primary modules: Spatial-Temporal Graph Encoder, Event Retrieval \& Text Embedding, Fusion \& Alignment, and Decoder. We first define the notations for traffic status, events, and the event aware traffic forecasting problem, followed by detailed explanations of Modules \textcircled{1}–\textcircled{4} in Sections \ref{sq:GE} - \ref{sq:dec}. A summary of the primary notations and their corresponding descriptions can be found in Table~\ref{Tb:Notationmodel}.

\subsection{Definitions}
\label{subsec:preliminary}
We first introduced the main definitions used in the paper.  
% \begin{definition}
% (Regions) We partition a city into N disjoint geographical grids, with each grid $v_n (1\leq n \leq N)$ representing a spatial region. We use $\mathcal{V}= \{v_1, ..., v_N\}$ to denote the region set.
% \end{definition}
\begin{definition}
[Road Network] We represent road network as an undirected graph $\mathcal{G}= \{\mathcal{V}, \mathbf{A}\}$, where $\mathcal{V}$ is the set of sensors and $\mathbf{A} \in \mathbb{R}^{N \times N}$ is a weighted adjacency matrix capturing
spatial dependencies between two sensors.
\end{definition}
We divide the entire time span into discrete intervals, denoted as $t = 1, ..., T$. 
\begin{definition}
[Traffic Status] Traffic Status (Traffic Flow/ Traffic Speed) can be represented as a two-dimensional tensor $\mathbf{X} = \{X_1, ..., X_T\}\in \mathbb{R}^{N\times T}$, where $T$ denotes the number of time intervals. Each element $X_t= (x^{(1)}_t, ..., x^{(N)}_t)$ ($t= 1, ..., T$) in the tensor denotes the traffic status values of all sensors at the $t$-th time interval.
\end{definition}
\begin{definition}
[Event] We use $C^s$ to denote the event feature, serving as the context for traffic status data, where $s \in \{1,2,...,N\}$ is the node id. Further, we use a tensor $\mathbf{C} = \{C^1, C^2,...,C^N\}$ to denote event features of all sensors.
\end{definition}
\begin{definition}
[Event Aware Traffic Forecasting] Based on the aforementioned definitions, the event aware traffic forecasting problem is to use traffic status data $X$ from past $H_{
\text{in}}$ timesteps and the event context $\mathbf{C}$ to forecast traffic flow data $\hat{Y}$ for future $H_{\text{out}}$ timesteps, which can be formulated as:
\begin{equation}
    \hat{Y}_{t + 1},..., \hat{Y}_{t + H_{\text{out}}} = f(X_{t - H_{\text{in}} + 1},..., X_t; \mathbf{C})
\end{equation}
where $f(\cdot)$ represents the prediction function.
\end{definition}
% Considering that traffic flow typically comprises continuous numerical values, in this paper, we treat traffic flow prediction as a regression task \cite{Lv-TITS2015}. 

\begin{table}[t]
\caption{Main notations and their descriptions}
\label{Tb:Notationmodel}
\centering
\resizebox{0.49\textwidth}{!}{%
\begin{tabular}{l p{4cm}}
\specialrule{1.5pt}{0pt}{0pt} 
\textbf{Symbol} & \textbf{Description} \\
\specialrule{1.5pt}{0pt}{0pt} 
$N$ & The total number of sensors \\
$T$ & The total length of the timesteps\\
$d$ & Embedding dimension \\
$x_t^{(s)}$ & Traffic status value of the $s$-th node \\
           & at the $t$-th time interval \\ 
$X_t= (x^{(1)}_t, ..., x^{(N)}_t)$ & Traffic status values of all $N$ sensors \\
           & at the $t$-th time interval. \\
$y^{(s)}_{t}$ & The ground truth traffic status for node $s$ at time $t$\\
% $\hat{y}^{(s)}_{t}/\Tilde{y}^{(s)}_{t}$ & The predicted flow of student/teacher \\
%            & model for region $s$ at time $t$\\
$H_{\text{in}}/H_{\text{out}}$ & The length of the input/output time window \\
$\mathbf{X} = \{X_1, ..., X_T\}\in \mathbb{R}^{N\times T}$ & Origin traffic status input\\
$\mathbf{Y}\in \mathbb{R}^{N\times T}$ & Ground truth traffic status\\
$\mathbf{\hat{Y}}\in \mathbb{R}^{N\times T}$ & Predicted traffic status\\
% $\mathbf{\tilde{Y}}\in \mathbb{R}^{N\times T}$ & Predicted flow of teacher model\\
$E_{\text{st}} \in \mathbb{R}^{N \times d}$ & Learnable spatial temporal embedding\\
$\mathbf{E}^{\text{text}} \in \mathbb{R}^{N \times d}$ & Learnable text embedding\\
$\mathbf{H}^{\text{fused}} \in \mathbb{R}^{N \times d}$ & Learnable Fused embedding\\
\hline
\end{tabular}%
}
\end{table}
\subsection{Spatial Temporal Graph Encoder}
\label{sq:GE}
The Spatial-Temporal Graph Encoder serves as the backbone for processing structured data within the FUSE-Traffic framework, as illustrated in Figure \ref{fig:arch} \textcircled{1}. Its primary role is to distill the complex, periodic patterns inherent in historical traffic data into a dense, informative embedding. To achieve this, we adopt D2STGNN\cite{shao2022decoupled} as the base graph encoder, chosen for its superior performance in decoupling and separately modeling spatial and temporal dependencies.
The resulting traffic embedding, which encapsulates these learned historical patterns, is denoted as $E_{\text{st}}$.
\subsection{Events Retrieval \& Embedding}
A core limitation in event-aware forecasting is the reliance on costly and often incomplete curated event datasets. Our framework addresses this limitation by introducing a dynamic, on-demand event retrieval module that leverages the advanced reasoning and extensive world knowledge of a Large Language Model (LLM). This approach, however, requires overcoming two main difficulties:
\begin{itemize}
    \item {High cost queries:} Spatial-temporal events typically occur over a wide area and last for a short period. This results in a large number of queries, many of which involve repeated events, making querying computationally expensive and inefficient.
    \item {Reasoning: } It is essential to guide the LLM in identifying which events could potentially impact traffic. This requires careful consideration of reasoning elements.
\end{itemize}

To address these challenges, for each recorded timestamp, we programmatically generate a prompt containing the geographic coordinates of all the sensors, as shown in Figure \ref{fig:arch} \textcircled{2}. The query period spans from $H_{\text{in}}$ to $H_{\text{out}} + H_{\text{in}} - 1$, ensuring that the events retrieved are both spatially and temporally relevant.

By considering local news, weather, crime and exhibitions, we enable LLM for retrieving dynamci events. Table~\ref{tab:COT} provides an example of this reasoning process and the formatted event output for a single sensor location and one specific observed timestamp.

% The [h!] option suggests LaTeX place the table "here" if possible.
% Adjust as needed (e.g., [htbp] for more float flexibility).
\begin{table}[h!]
\centering % Centers the table on the page
\caption{LLM response example: location (34.0522° N, -118.2437° W), timestamp '2012-03-02 17:40'}
\label{tab:COT}
\begin{tabularx}{0.48\textwidth}{lX} % 'l' for the first column (left-aligned, natural width)
                                     % 'X' for the second column (expandable, wraps text)
\toprule
\textbf{Reasoning Element} & \textbf{LLM Response} \\
\midrule
Sensor Location & Downtown Los Angeles \\
\midrule
Event Time Window& Start: 18:40, End: 19:35 (Friday Evening) \\
\midrule
Weather & Partly Cloudy \\
\midrule
Identified Event(s): Sport & LA Lakers Game at Staples Center (approx. 7:30 PM start) \\
\midrule
Identified Event(s): Exhibition & Wilco concert at The Wiltern upcoming/starting \\
\midrule
Synthesized Output & \{"Event": "LA Lakers Game at Staples Center (approx. 7:30 PM start), Typical Friday evening traffic patterns, Wilco concert at The Wiltern upcoming/starting"\} \\
\bottomrule
\end{tabularx}
\end{table}

The event texts extracted from the LLM's JSON responses for all sensors, collectively forming the event feature $C$, are then converted into a dense vector representation $E_{\text{text}}$, using the pretrained text encoder. We freeze the text encoder parameters in the training stage to
keep the powerful generalization ability learned from massive text
training data for sparse event text embedding.
\subsection{Fusion \& Alignment}
\label{sq:fu}
For our event aware traffic forecasting task, the input comprise two modalities, spatial-temporal (ST) traffic status data and textual event data. After processing these modalities with their respective encoders (e.g., a graph encoder for ST data and a text encoder for event data), latent representations, denoted as $E_{\text{st}}$ and $E_{\text{text}}$, are obtained. To effectively leverage the complementary information from these heterogeneous sources, we need a method to fuse and align the two modalities. 
% The common used mechanisms and method choosen by FUSE-Traffic are illustrated in Figure \ref{fig:fusion}.
% \begin{figure}
%     \centering
%     \includegraphics[width=1\linewidth]{fig/fusion.pdf}
%     \caption{Overview of common fusion and alignment mechanism. The method proposed by FUSE-Traffic is highlighted in red and bottom.}
%     \label{fig:fusion}
% \end{figure}

To aggregate the time series and the text modalities, we design a cross-modality fusion and alignment block, as shown in Figure \ref{fig:arch} \textcircled{3}. Spatial Temporal Graph embeddings $E_{\text{st}}$ from Graph Encoder encode highlevel temporal patterns and serve as queries, while the text embeddings $E_{\text{text}}$ from the VLM serve as keys and
values.
The multi-head attention is defined as:
\begin{equation}
    MHA(Q, K, V ) = Concat(head1, . . . , headh)W^O
\end{equation}
\begin{equation}
    head_i = \text{softmax}(\frac{QW^Q(KW^K)^T}{\sqrt{d_k}})VW^V
\end{equation}

where $Q = E_{\text{st}}W^Q, K = E_{\text{text}}W^K, V =  E_{\text{text}}W^V$. Here, $W^Q, W^K, W^V$ are all learnable parameters. $d_k = \frac{d}h$ is the head dimension, and $h$ is the number of attention heads. 

Then we perform layer normalization $LN (\cdot)$ before adding the $E_{\text{st}}$ to the output of the multi-head attention:
\begin{equation}
    H_C = LN(MHA(Q,K,V) + E_{\text{st}})
\end{equation}

Next, we pass $H_C$ through a feed-forward network (FFN) followed by another layer normalization to obtain $H_{\text{fused}}$:

\begin{equation}
    H_{\text{fused}} = LN(H_C + FFN(H_C))
\end{equation}
where $FFN(\cdot)$ denotes feed-forward layer.

This mechanism effectively aligns and fuses both spatial-temporal and event text features, enabling the model to capture both fine-grained patterns and high-level context.

\subsection{Decoder}
\label{sq:dec}
Finally, the $H_{\text{fused}}$is input into a fully connected layer $FC(\cdot)$ for
future prediction, which is formulated as follows:
\begin{equation}
\hat{\mathbf{Y}} = FC(H_{\text{fused}})
\end{equation}

\section{Peformance Evaluation}
We investigate the effectiveness of our model with the
goal of answering the following research questions:
\begin{itemize}
    \item RQ1: Does our FUSE-Traffic outperform other baselines?
    \item RQ2: Can the proposed model robustly handle the predicting tasks with varying traffic patterns?
    \item RQ3: How do hyper-parameters affect FUSE-Traffic?
    \item RQ4: How do framework and components in FUSE-Traffic(e.g., prompt, fusion mechanisms) affect model performance?
    % \item RQ4: Does our FUSE-Traffic efficient and effective?
    \item RQ5: How does Event Knowledge guide FUSE-Traffic in Traffic Forecasting?
\end{itemize}

\subsection{Settings}
\subsubsection{Dataset Description}
To thoroughly evaluate the performance of our proposed method, we conduct experiments using three real-world traffic datasets, covering traffic speed and traffic flow: 
\begin{itemize}
    \item METR-LA\cite{li2017diffusion}: This dataset contains traffic speed time series recorded by 207 sensors on highways in Los Angeles County, USA. The traffic information was collected every 5 minutes, resulting in a total of 34,272 time steps.
    \item PEMS-BAY\cite{li2017diffusion}: This dataset is a traffic speed time series dataset recorded by sensors at 325 different locations and collected by the California Transportation Agencies (CalTrans) Performance Measurement System (PeMS). The data is collected every 5 minutes, spanning 52,116 time steps.
    \item PEMS03\cite{Song_Lin_Guo_Wan_2020}: This is a traffic flow dataset also collected from CalTrans PeMS. The data covers a 2-month period from September 1, 2018, to November 30, 2018, with a total of 26,208 time steps.
\end{itemize}

For data preprocessing, we apply Z-score normalization to standardize the raw input values from all datasets. A statistical overview of these datasets is provided in Table \ref{tab:dataset_statistic}.
\begin{table}[h!]
    \centering
    \caption{The statistic of datasets: Time Span (mm/dd/yy)}
    \label{tab:dataset_statistic}
        \resizebox{\linewidth}{!}{ % This line resizes the table to the width of the frame
    \begin{tabular}{lccccc}
        \specialrule{1.5pt}{0pt}{0pt} 
        Dataset & Type&\# Sensors & Sample Rate & Time Span\\ \hline
        METR-LA & Traffic Speed &207 & 5 mins & 03/01/12 – 06/27/12 \\
        PEMS-BAY & Traffic Speed &325 & 5 mins & 01/01/17 – 06/30/17\\
        PEMS03 & Traffic Flow &358 & 5 mins &  09/01/18 - 11/30/18\\
        % DiDi-BJ(Not public) &  99,716 & 5mins & 01/01/21 - 01/31/21 \\
        \specialrule{1.0pt}{0pt}{0pt} 
    \end{tabular}
    }
\end{table}
\subsubsection{Evaluation Metrics}

We evaluate the accuracy of the traffic flow prediction using \emph{Mean Absolute Error (MAE)}, \emph{Root Mean Square Error (RMSE)} and Mean Absolute Percentage Error (MAPE). These metrics allow us to measure the relative error of the estimated inflow and outflow. They are defined as:
\begin{equation}
\label{eq:MAE}
\text{MAE}(\mathbf{Y}, \hat{\mathbf{Y}}) = \frac{1}{N \times T} \sum_{i=1}^{N \times T} | \hat{y}_i - y_i |
\end{equation}
\begin{equation}
\text{RMSE}(\mathbf{Y}, \hat{\mathbf{Y}}) = \sqrt{\frac{1}{N \times T} \sum_{i=1}^{N \times T} ( \hat{y}_i - y_i )^2 }
\end{equation}
\begin{equation}
\label{eq:MAPE}
\text{MAPE}(\mathbf{Y}, \hat{\mathbf{Y}}) = \frac{1}{N \times T} \sum_{i=1}^{N \times T} \left| \frac{\hat{y}_i - y_i}{y_i} \right| \times 100\%
\end{equation}
where $y$ represents the actual inflows/outflows, $\hat{y}$ is the predicted inflows/outflows, $N$ is the total number of sensors, and $T$ is the total number of time intervals. 
\subsubsection{Baselines}
We consider 8 baseline models, categorized into four categories, all of which have demonstrated strong performance in traffic prediction tasks. The taxonomy of these traffic forecasting methods, with regard to modality and whether they use fusion/alignment, is illustrated in Table~\ref{tab:tax}.
\begin{itemize}
    
\item [(1)] \textbf{GNN-based Models(w/o events)}
\begin{itemize}
\item GraphWaveNet \cite{Wu2019GraphWaveNet}: It integrates graph convolutional networks (GCN) with dilated causal convolutions to capture both spatial dependencies and long-range temporal patterns.
\item STWave \cite{Fang2023STWave}: This approach utilizes a discrete wavelet transform to decompose traffic time series into multiple frequency components. 
\item D2STGNN \cite{shao2022decoupled}:  It proposes a decoupled dynamic spatial-temporal graph neural network that separately models spatial and temporal dependencies while accounting for dynamic correlations among traffic sensors.
\end{itemize}

\item [(2)] \textbf{GNN-based Models(Manually Engineered Event Features):}
\begin{itemize}
    \item DIGC-Net \cite{Xie2020DIGCNet}: It introduces a method to identify "critical incidents" that significantly impact traffic flow and employs a binary classifier to extract latent features representing the impact of these incidents.
\end{itemize}
\item [(3)] \textbf{GNN-based Models with Event Semantics}
\begin{itemize}
    \item STORM \cite{Hong2024STORM}: It models event impacts using two distinct fusion diagrams: a multi-task one with an attention mechanism for temporal effects and a multi-view one with a stimulus-response mechanism for spatial effects.
    \item T3 \cite{EventTraffic2024}:A multi-modal event traffic forecasting model that uses pre-trained text and traffic encoders to extract the embeddings and fuses the two embeddings for prediction.
\end{itemize}
\item [(4)] \textbf{LLM-based Models:} 
\begin{itemize}
    \item UrbanGPT \cite{Li2024UrbanGPT}: A spatio-temporal large language model that integrates a specialized dependency encoder with an instruction-tuning paradigm, enabling it to understand complex spatio-temporal patterns
    \item STD-PLM \cite{huang2025stdplmunderstandingspatialtemporal}: It adapts Pre-trained Language Models (PLMs) for spatio-temporal tasks by using specialized spatial and temporal tokenizers.
\end{itemize}
\end{itemize}

\begin{table}[h!]
    \centering
    \small
    \setlength{\tabcolsep}{1pt} % Adjust column separation for better 
    \caption{Taxonomy of Traffic Forecasting Methods. Modality Refers to the different data modalities involved in each method.}
    \label{tab:tax}
        \resizebox{\linewidth}{!}{ % This line resizes the table to the width of the frame
    \begin{tabular}{lcccc}
        \specialrule{1.5pt}{0pt}{0pt} 
        Method & Modality & Fusion & Alignment & Backbone \\ \hline
        GraphWaveNet\cite{Wu2019GraphWaveNet} & Time Series & $\times$& $\times$& GNN\\
        STWave\cite{Fang2023STWave}  & Time Series& $\times$& $\times$& GNN\\
        D2STGNN \cite{shao2022decoupled} & Time Series& $\times$& $\times$& GNN\\
        DIGCNet\cite{Xie2020DIGCNet} & Time Series& $\times$& $\times$& GNN\\
        STORM\cite{Hong2024STORM} & Time Series \& Text & Intermediate:Add & $\checkmark$(gating) & GNN\\
        T3\cite{EventTraffic2024} & Time Series \& Text & Output & $\times$& GNN, T3\\
        UrbanGPT\cite{EventTraffic2024} & Time Series \& Text & Input & $\times$& LLM\\
        STD-PLM\cite{huang2025stdplmunderstandingspatialtemporal} & Time Series & $\times$ & $\times$& LLM\\
        FUSE-Traffic & Time Series \& Text & Intermediate:Add & $\checkmark$(cross attention) & GNN, LLM\\
        \specialrule{1.0pt}{0pt}{0pt} 
    \end{tabular}
    }
\end{table}
\begin{figure}[htbp]
    % \centering
    \includegraphics[width=1\linewidth]{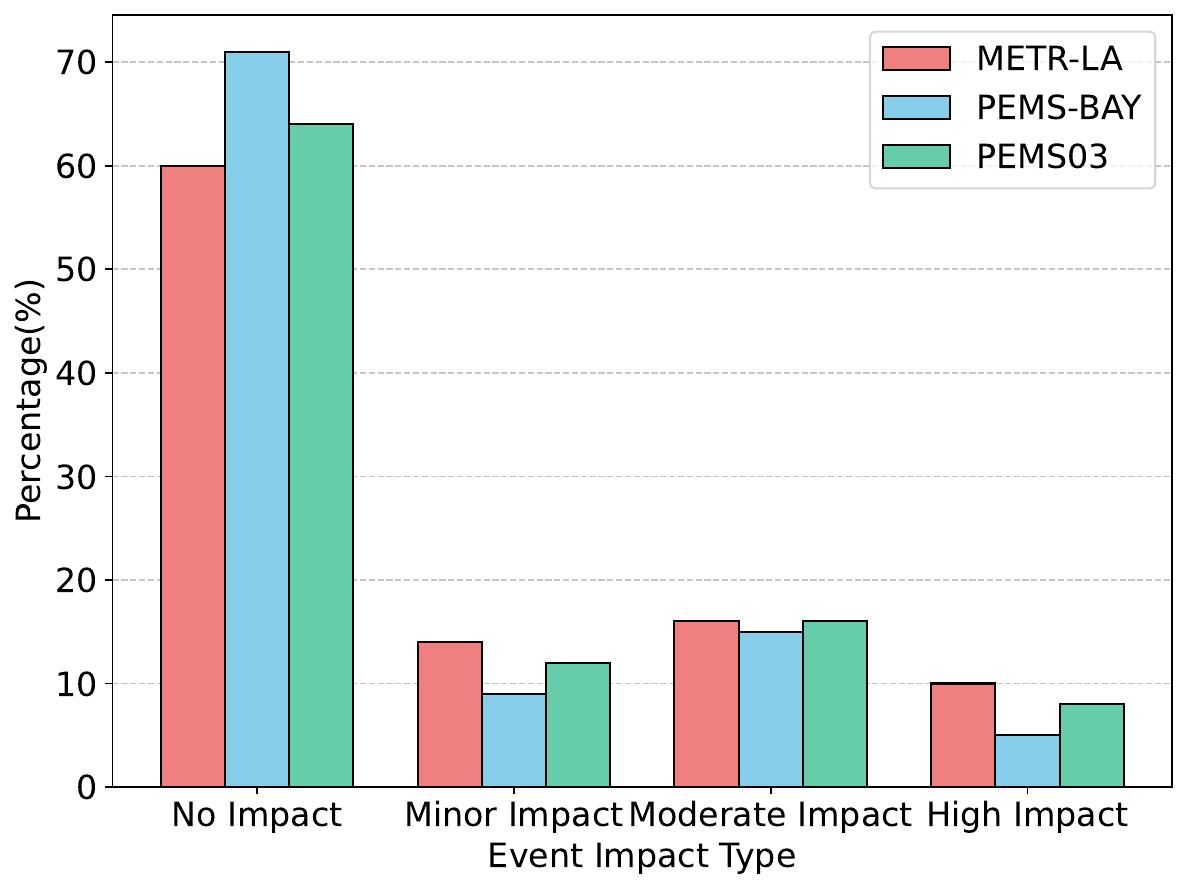}
    \caption{Distribution of Event Impact Type in 3 Datasets}
    \label{fig:distribution}
\end{figure}
\begin{table*}[htbp]
\centering
\caption{Performance Comparison on METR-LA and PEMS-BAY. Best and second-best results are indicated in \textbf{bold} and \underline{underlined}, respectively.}
\label{table:metrla_pemsbay_reformatted}
% \footnotesize % Use a smaller font size for large tables
\setlength{\tabcolsep}{2.5pt} % Adjust column spacing
\begin{tabular}{l|c |ccc| ccc |ccc| ccc}
\toprule
\multirow{2}{*}{\textbf{Datasets}} & \multirow{2}{*}{\textbf{Methods}} & \multicolumn{3}{c|}{\textbf{Horizon 3 (15 min)}} & \multicolumn{3}{c|}{\textbf{Horizon 6 (30 min)}} & \multicolumn{3}{c|}{\textbf{Horizon 12 (60 min)}} & \multicolumn{3}{c}{\textbf{Average}} \\
\cmidrule(lr){3-5} \cmidrule(lr){6-8} \cmidrule(lr){9-11} \cmidrule(lr){12-14}
& & MAE & RMSE & MAPE(\%) & MAE & RMSE & MAPE(\%) & MAE & RMSE & MAPE(\%) & MAE & RMSE & MAPE(\%) \\
\midrule
\multirow{9}{*}{METR-LA}
& GraphWaveNet & 2.70 & 5.18 & 7.21 & 3.09 & 6.25 & 9.33 & 3.52 & 7.34 & 10.49 & 3.10 & 6.26 & 9.01 \\
& STWave & 2.83 & 5.65 & 8.08 & 3.22 & 6.73 & 10.02 & 3.55 & 7.57 & 11.23 & 3.20 & 6.65 & 9.78 \\
& D2STGNN & 2.57 & \underline{4.96} & \underline{7.12} & 2.99 & \underline{6.02} & 9.44 & 3.42 & \underline{7.31} & 9.99 & 3.00 & \underline{6.10} & 8.85 \\
& DIGC-Net & 2.88 & 5.59 & 8.22 & 3.18 & 6.65 & 10.72 & 3.49 & 7.49 & 10.12 & 3.19 & 6.58 & 9.69 \\
& STORM & 2.64 & 4.96 & 7.14 & 3.00 & 6.13 & \underline{8.11} & 3.45 & 7.69 & 10.04 & 3.03 & 6.26 & \underline{8.43} \\
& T3 & \underline{2.56} & 5.01 & 8.33 & \underline{2.98} & 6.04 & 9.46 & 3.46 &7.39 & \underline{9.96} & \underline{3.00} & 6.15 & 9.25 \\
& UrbanGPT & 3.02 & 5.88 & 11.01 & 3.28 & 6.77 & 13.01 & 3.50 & 7.60 & 10.21 & 3.27 & 6.75 & 11.41 \\
& STD-PLM & 2.62 & 5.02 & 10.22 & 3.05 & 6.12 & 9.67 & \textbf{3.38} & 7.36 & 9.98 & 3.02 & 6.17 & 9.96 \\
& \textbf{FUSE-Traffic} & \textbf{2.53} & \textbf{4.93} & \textbf{6.54} & \textbf{2.90} & \textbf{5.89} & \textbf{8.02} & \underline{3.39} & \textbf{7.28} & \textbf{9.84} & \textbf{2.94} & \textbf{6.03} & \textbf{8.13} \\
\midrule
\multirow{9}{*}{PEMS-BAY}
& GraphWaveNet & 1.31 & 2.75 & 3.91 & 1.66 & 3.78 & 4.56 & 2.00 & 4.67 & 5.11 & 1.66 & 3.73 & \underline{4.53} \\
& STWave & 1.37 & 2.74 & 4.12 & \underline{1.64} & \underline{3.72} & 5.12 & 1.91 & 4.37 & 5.03 & 1.64 & \underline{3.61} & 4.76 \\
& D2STGNN & 1.34 & 2.80 & 3.94 & 1.69 & 3.84 & 4.87 & 2.00 & 4.54 & 5.24 & 1.68 & 3.73 & 4.68 \\
& DIGC-Net & 1.32 & 2.79 & 3.86 & 1.68 & 3.86 & 5.32 & 1.94 & 4.46 & 5.19 & 1.65 & 3.70 & 4.79 \\
& STORM & \underline{1.28} & \underline{2.71} & \underline{3.77} & 1.66 & 3.79 & 4.99 & 1.97 & 4.53 & 6.12 & 1.64 & 3.67 & 4.96 \\
& T3 & 1.34 & 2.79 & 4.26 & 1.65 & 3.73 & \underline{4.52} & 1.90 & 4.35 & 4.99 & \underline{1.63} & 3.62 & 4.59 \\
& UrbanGPT & 1.47 & 2.85 & 5.01 & 1.69 & 3.78 & 6.00 & 2.00 & 4.69 & 6.05 & 1.72 & 3.77 & 5.69 \\
& STD-PLM & 1.42 & 2.82 & 3.88 & 1.68 & 3.77 & 5.01 & \textbf{1.85} & \textbf{4.32} & \textbf{4.88} & 1.65 & 3.64 & 4.59 \\
& \textbf{FUSE-Traffic} & \textbf{1.27} & \textbf{2.68} & \textbf{3.71} & \textbf{1.59} & \textbf{3.63} & \textbf{4.48} & \underline{1.88} & \underline{4.34} & \underline{4.91} & \textbf{1.58} & \textbf{3.55} & \textbf{4.37} \\
\midrule
\multirow{9}{*}{PEMS03}
& GraphWaveNet & 14.55 & 24.60 & 15.02 & \underline{15.59} & 26.70 & \underline{15.70} & \underline{17.28} & \underline{29.28} & \underline{16.84} & \underline{15.81}&	\underline{26.86}	&\underline{15.85} \\
& STWave & 14.69 & 25.67 & 15.26 & 15.82 & 27.58 & 16.16 & 17.46 & 29.87 & 17.69 & 15.99	&27.71	&16.37 \\
& D2STGNN & \underline{14.40} & \underline{24.34} & \underline{14.78} & 15.62 & \underline{26.52} & 15.73 & 17.59 & 30.16 & 17.63 & 15.87&	27.01	&16.05 \\
& DIGC-Net & 14.53 & 24.71 & 14.93 & 16.26 & 27.46 & 16.48 & 19.29 & 31.79 & 18.77 & 16.69	&27.99&	16.73\\
& STORM & 15.25 & 25.87 & 18.96 & 16.73 & 28.25 & 21.30 & 19.10 & 31.74 & 21.77 & 17.03&	28.62&	20.68\\
& T3 & 14.80 & 24.65 & 22.26 & 16.31 & 27.16 & 24.29 & 18.73 & 31.02 & 25.20 & 16.61	&27.61&	23.92\\
& UrbanGPT & 15.62 & 25.80 & 23.21 & 17.52 & 28.87 & 23.73 & 21.36 & 34.55 & 25.08 & 	18.17	&29.74&	24.01\\
& STD-PLM & 14.49 & 24.68 & 14.89 & 16.09 & 27.35 & 15.74 & 18.82 & 31.52 & 18.91 & 16.47	&27.85&	16.51 \\
& \textbf{FUSE-Traffic} & \textbf{13.57} & \textbf{24.11} & \textbf{14.44} & \textbf{14.83} & \textbf{26.44} & \textbf{15.24} & \textbf{16.75} & \textbf{29.17} & \textbf{16.15} & \textbf{15.05}	&\textbf{26.57}&	\textbf{15.28}\\
\bottomrule
\end{tabular}
\end{table*}
\subsubsection{Experiment Settings}
All experiments were conducted on a Linux server equipped with four NVIDIA A6000 GPUs, each with 48 GB of memory. For the METR-LA and PEMS-BAY datasets, the data was partitioned into training (70\%), validation (10\%), and testing (20\%) sets. For the PEMS03 dataset, a 60\%/20\%/20\% split was used for training, validation, and testing, respectively. Since the events retrieved by the LLM include traffic impact information, we categorize the events into four types: No Impact, Minor Impact, Moderate Impact, and High Impact. The distribution of these event types across the three datasets is shown in Figure \ref{fig:distribution}. We use D2STGNN \cite{shao2022decoupled} as the base graph encoder with 5 decoupled Spatial Temporal Layer. The LLM and text encoder used is Gemini 2.5 pro review and Gemini text-embedding-005. To ensure architectural compatibility for the subsequent fusion stage, both the resulting spatio-temporal and text embeddings are projected to a unified dimension of 1024.

To ensure reproducibility, the key hyperparameters for each baseline are detailed below, based on their official implementations. Early stopping with a patience of 10 epochs was used across all experiments to prevent overfitting. 
% Below is a summary of the critical configurations for each baseline:
\begin{itemize}
    \item {\bf GraphWaveNet}: The GraphWaveNet model includes 2 graph wavelet layers
and 4 blocks. The hidden dimension is 512, and a dropout rate of 0.3 is used. The training is performed with a learning rate of 0.002, a batch size of 128, and 100 epochs.
    \item {\bf STWave:} The STWave model consists of 2 layers of spatial temporal encoder with a hidden dimension of 128, the sampling factor of ESGAT is set to 1. The model is trained with a learning rate of 0.002, a batch size of 128, and 100 epochs.
    \item {\bf D2STGNN:} The D2STGNN model comprises 5 decoupled Spatial Temporal Layer. The hidden dimension is set to 32, and a dropout rate of 0.5 is applied. The model is trained with a learning rate of 0.002. a batch size of 128, and 100 epochs.
    \item {\bf DIGCNet:} The DIGCNet model consists of two GCN layers, followed by a fully connected layer. For the weather and incident dataset, we retain the same information as in our model. For context learning, since LLM does not provide details such as the start and end or the duration of an incident, we only consider the incident type and road status as context information. We classify high-impact events as critical incidents, while all other events (excluding those with no impact) are categorized as non-critical incidents. The model is trained with a learning rate of 0.001, a batch size of 128, and 100 epochs.
    \item {\bf STORM:} The STORM model includes 1 GCN layer and 1 GRU layer. The hidden dimension is 64. For the meteorological data and social event data, we retain the same information as in our model. The model is trained with a learning rate of 0.002, a batch size of 128, and 100 epochs.
    \item {\bf T3:} The T3 model uses GraphWaveNet as the base traffic encoder and Voyage-2 as the text encoder. To ensure a fair comparison, we retain the same event information as in our model. The model is trained with a learning rate of 0.0002, a batch size of 64 and 100 epochs.
    \item {\bf UrbanGPT:} The UrbanGPT model involves pretraining of the spatial-temporal dependency encoder and intruction-tuning of LLM. To ensure a fair comparison, we perform the instruction tuning solely on the traffic dataset, rather than on multiple datasets (Bike, Crime, Flow) as used in the original paper. The model is trained with a learning rate of 0.002, a batch size of 4, and 3 epochs.
    \item {\bf STD-PLM:} The STD-PLM model is implemented with 3 PLM layers, each with word embedding dimension of 768. A dropout rate of 0.1 is applied, and the model is trained with a learning rate of 0.001, a batch size of 64 and 500 epochs.
\end{itemize}
\subsection{Performance Comparison (RQ1)}

To evaluate the performance and generalization capability of our proposed FUSE-Traffic model, we conducted a comprehensive comparison against eight baseline models on three public benchmark datasets: METR-LA, PEMS-BAY, and PEMS03. The detailed results for short, medium, and long-term forecasting horizons (15, 30, and 60 minutes) are presented in Table \ref{table:metrla_pemsbay_reformatted}.
%The results clearly demonstrate that 

From the results, we obtain the
following observations: (1) Our proposed model, FUSE-Traffic outperforms all baselines across almost all forecasting horizons and consistently achieves the best average performance on all three datasets. This consistent superiority underscores the effectiveness of our proposed method, which jointly learns from spatio-temporal traffic data and external event knowledge. (2) Compared with other fusion-based models STORM,
T3, and UrbanGPT, Our model achieves better performance, highlighting the superiority of our fusion and alignment scheme.
\subsection{Robustness Analysis (RQ2)}
\begin{figure}[htbp]
    % \centering
    \includegraphics[width=\linewidth]{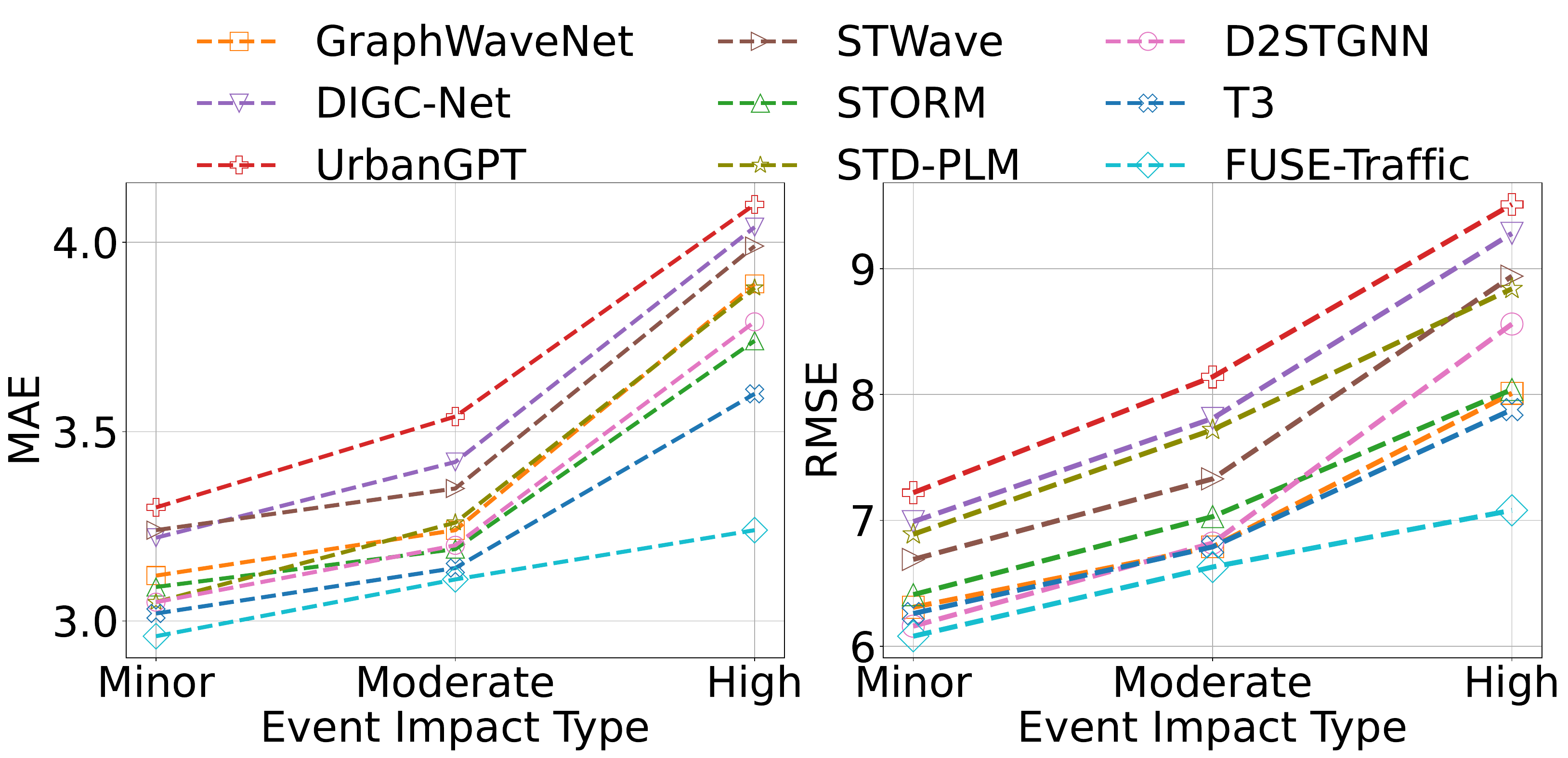}
    \caption{Performance Comparison in METR-LA}
    \label{fig:raMT}
\end{figure}

\begin{figure}[htbp]
    \includegraphics[width=\linewidth]{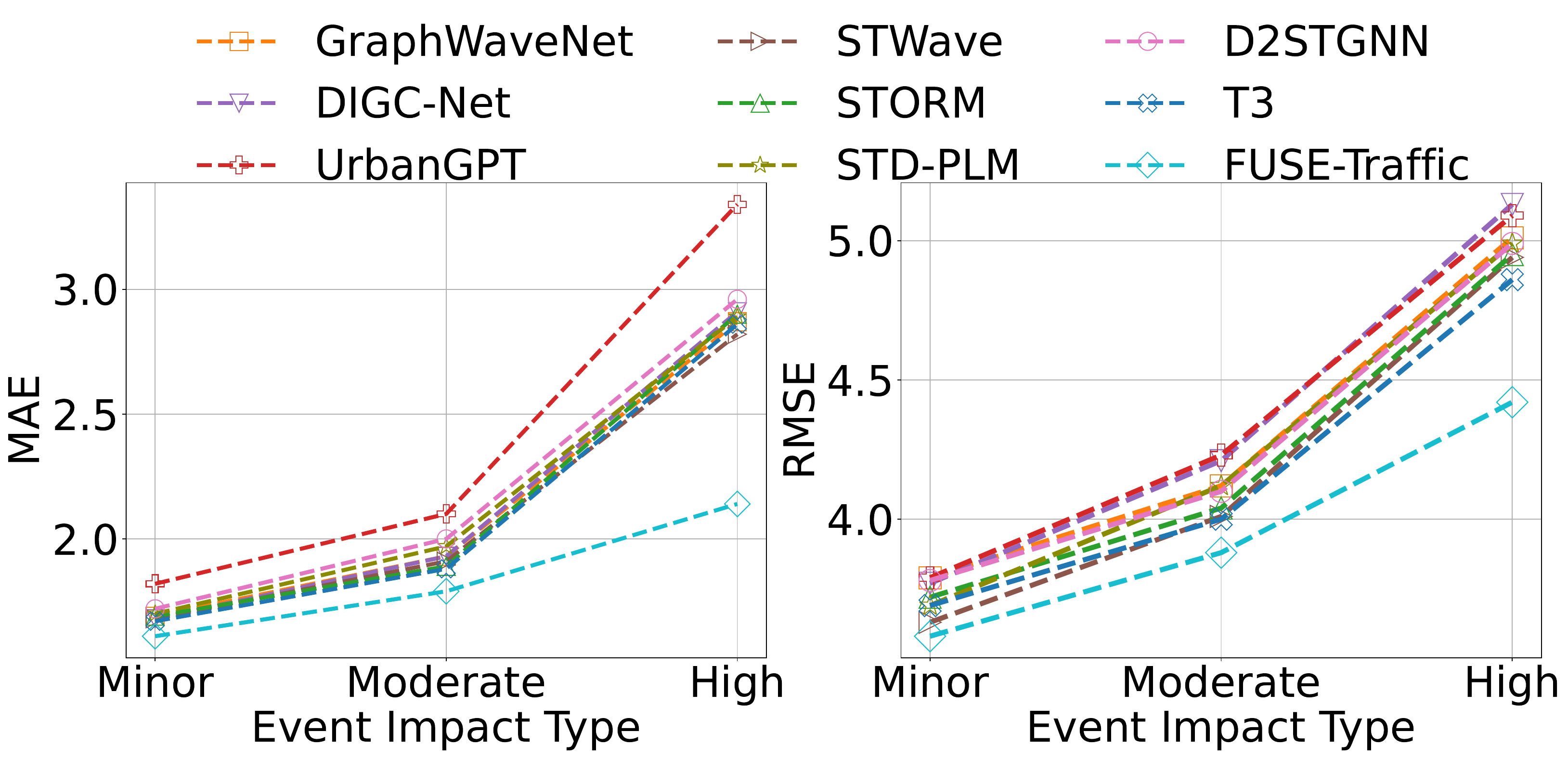}
    \caption{Performance Comparison in PEMS-BAY}
    \label{fig:raBAY}
\end{figure}

\begin{figure}[htbp]
    \includegraphics[width=\linewidth]{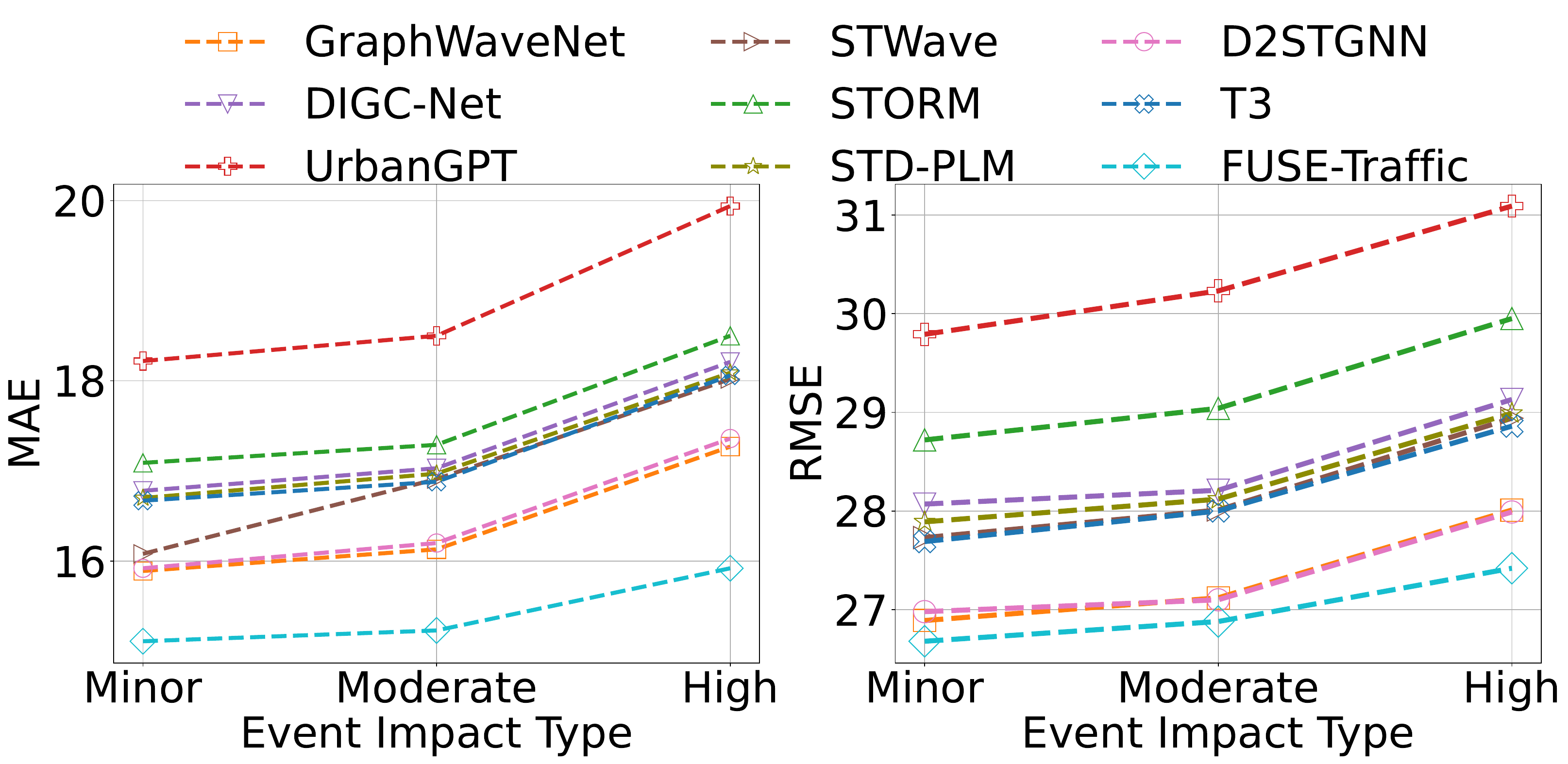}
    \caption{Performance Comparison in PEMS03}
    \label{fig:raPS}
\end{figure}

We assess the performance of the models across various event impact scenarios in three traffic datasets, as depicted in Figures \ref{fig:raMT} - \ref{fig:raPS}. These figures clearly demonstrate the superior performance of our model (highlighted in cyan). Moreover, as the event impact increases, our model consistently outperforms the baseline models, showing a significant advantage in terms of MAE and RMSE.


\subsection{Hyper-parameter Study (RQ3)}

We conduct experiments to analyze the impacts of two hyperparameters: the number of MLP layer $l$ and the text encoder embedding dimension. We present the results on PEMS-BAY dataset. As can be seen, when adjusting the number of layers and the text encoder embedding size, the model performance first improves and then stabilizes. This can be attributed to the increased expressive power as model parameter grows. However,
due to the limited data volume and diversity, expanding the model
parameters may eventually hit a performance bottleneck. The model hits the best performance under MLP layer 2 and text encoder embedding size 512.

\begin{figure}[htbp]
    \centering
    % Create the first mini-page for the left image
    \begin{minipage}{0.49\linewidth}
        \centering
        \includegraphics[width=\linewidth]{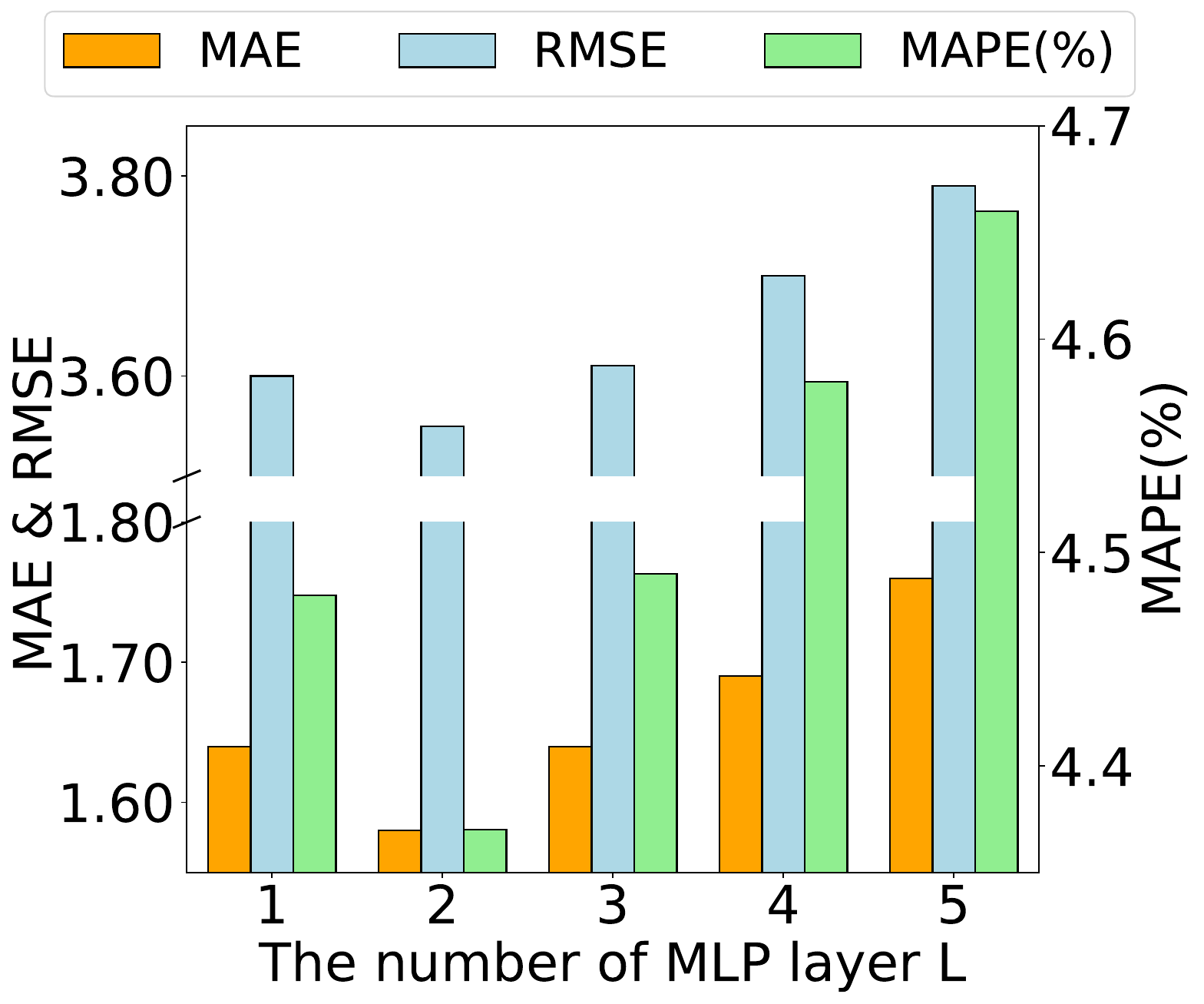}
    \end{minipage}
    \hfill % This command creates flexible horizontal space between the images
    % Create the second mini-page for the right image
    \begin{minipage}{0.49\linewidth}
        \centering
        \includegraphics[width=\linewidth]{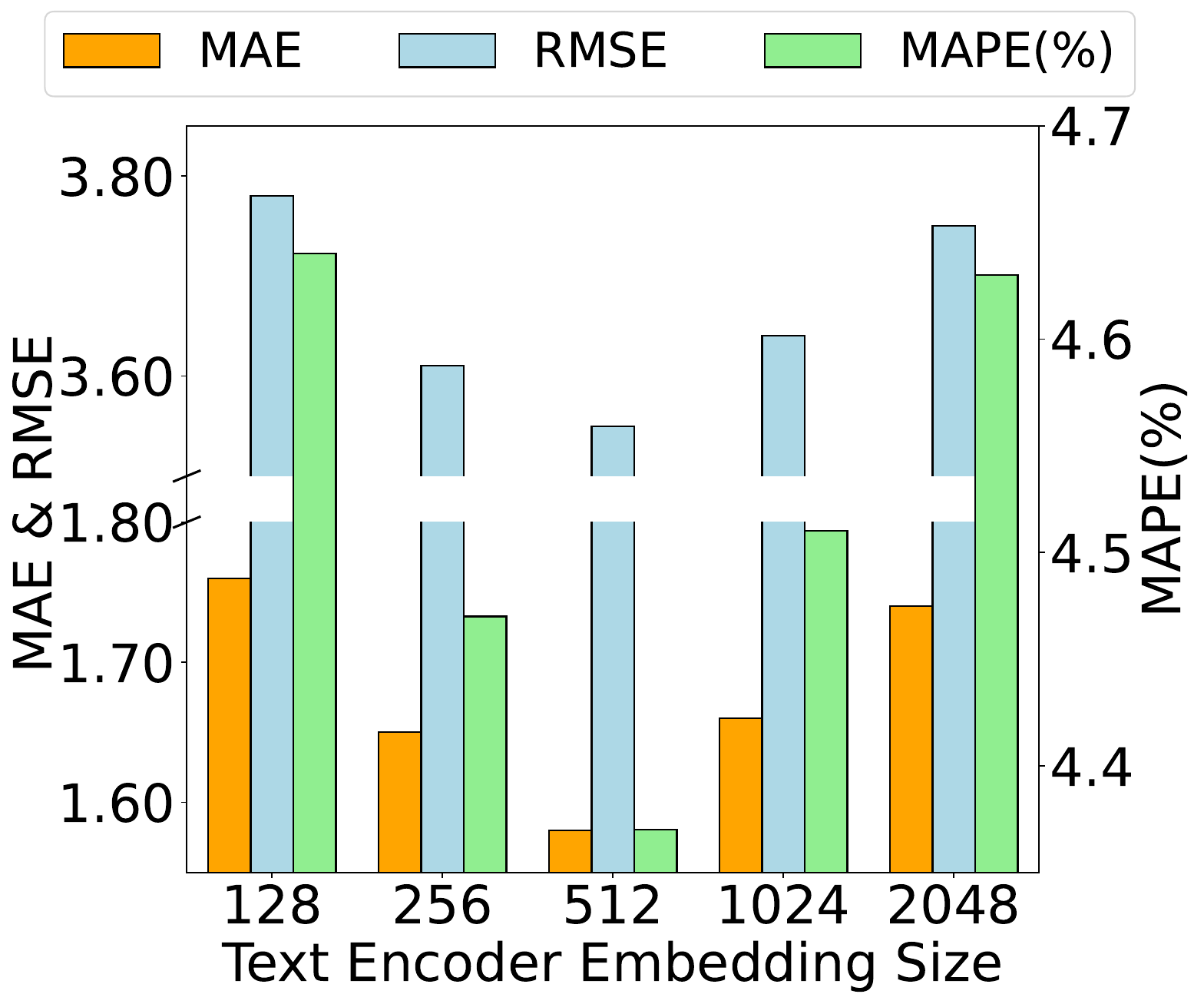}
        % No caption or label here
    \end{minipage}
    
    % The single, shared caption for the entire figure
    \caption{Hyperparameter sensitivity experiment in PEMS-BAY}
    \label{fig:hyper_combined}
\end{figure}
\subsection{Ablation Study (RQ4)}
In this section, we perform ablation studies on variants of model design and prompt design.

\begin{table}[h!]
    \centering
    \small
    \setlength{\tabcolsep}{2pt} % Adjust column separation
    \caption{Prompt Design}
    \label{tab:prompt}
    \begin{tabularx}{\linewidth}{lX} % Use tabularx for width control and X column for wrapping
        \specialrule{1.5pt}{0pt}{0pt} % Consider using \toprule from booktabs package
        Prompt Index & Prompt Content \\ \hline
        P1(Full Feature) & For each sensor, identify nearby events that could impact traffic (e.g., LA news, Severe Weather, concerts, crime). Example: Classic Cinema Night at Cinegrill Theater.\\
        P2(Single Event) & For each sensor, identify the most influential event that could impact traffic (e.g., LA news, Severe Weather, concerts, crime). Example: Classic Cinema Night at Cinegrill Theater.\\
        P3(w/o considering weather) & For each sensor, identify nearby events that could impact traffic (e.g., LA news, concerts, crime). Example: Classic Cinema Night at Cinegrill Theater.\\ 
        P4(w/o considering crime) & For each sensor, identify nearby events that could impact traffic (e.g., LA news, Severe Weather, concerts). Example: Classic Cinema Night at Cinegrill Theater.\\ 
        P5(Zero Shot) & For each sensor, identify nearby events that could impact traffic (e.g., LA news, Severe Weather, concerts).  \\
        \specialrule{1.0pt}{0pt}{0pt} % Consider using \bottomrule from booktabs package
    \end{tabularx}
\end{table}
\begin{figure}
    \centering
    \begin{minipage}{0.49\linewidth}
        \centering
        \includegraphics[width=\linewidth]{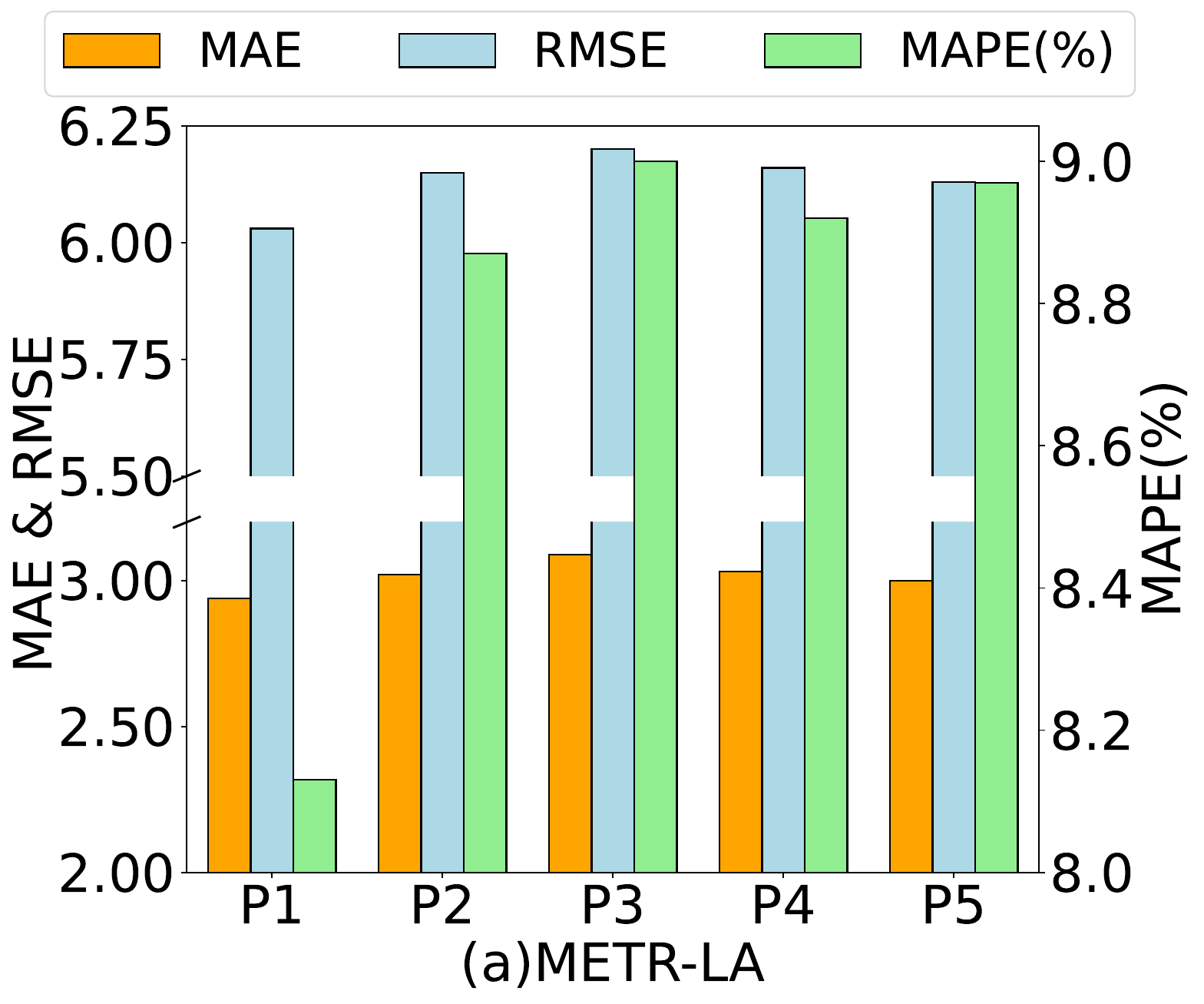}
    \end{minipage}
    \hfill % This command creates flexible horizontal space between the images
    % Create the second mini-page for the right image
    \begin{minipage}{0.49\linewidth}
        \centering
        \includegraphics[width=\linewidth]{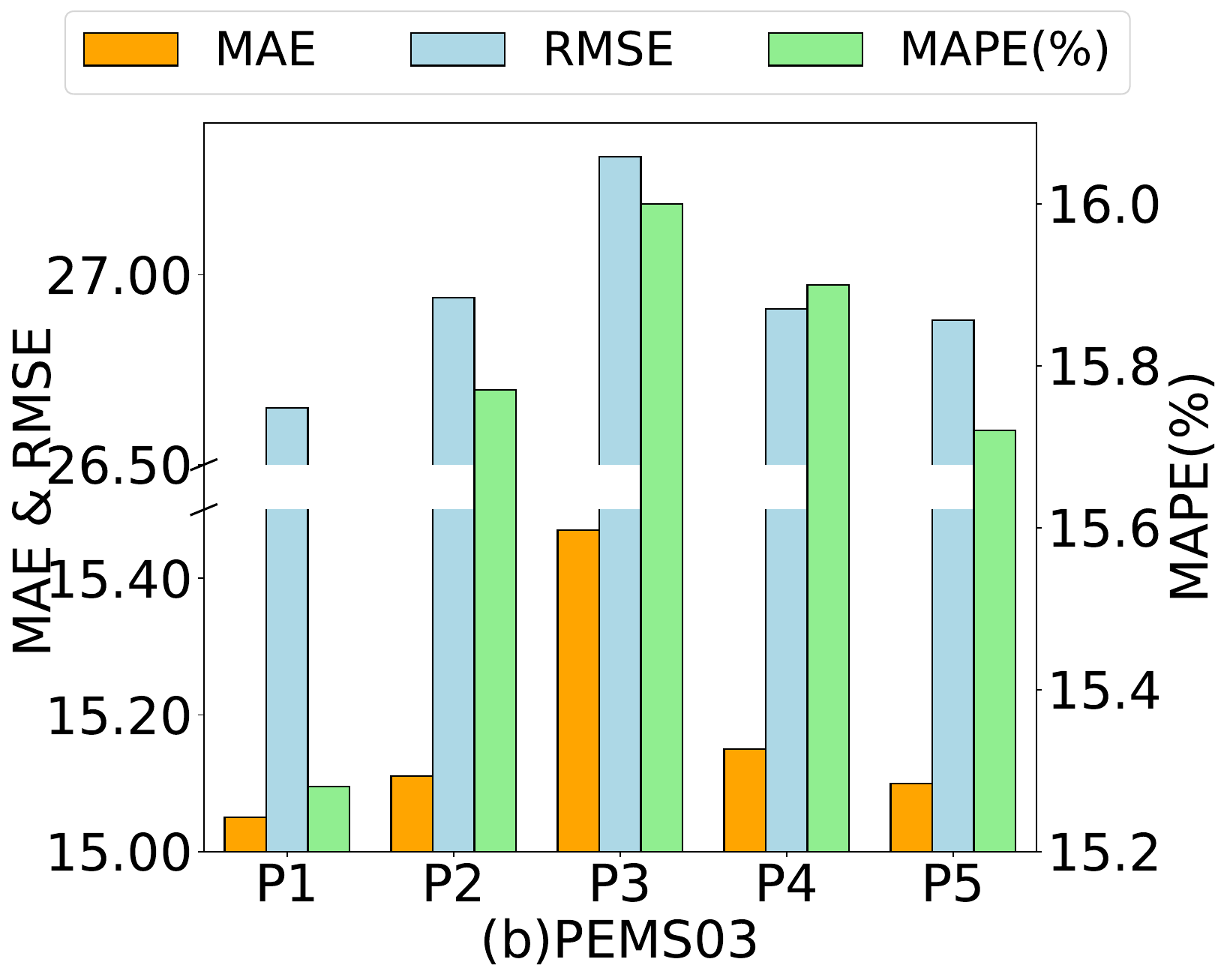}
        % No caption or label here
    \end{minipage}
    \caption{Ablation Study on Prompt Design}
    \label{fig:Prompt}
\end{figure}

\subsubsection{Prompt Design}
We design five distinct prompts, detailed in Table \ref{tab:prompt}. Specifically, Prompt P1 retains the full feature set. Prompt P2 focuses solely on retrieving the most influential events that could impact traffic. In Prompt P3, we do not explicitly instruct the LLM to consider weather as a type of event. Prompt P4 excludes crime incidents, and in Prompt P5, we refrain from providing the LLM with event examples. The results are presented in Figure \ref{fig:Prompt}. Firstly, we observe significant performance degradation w/o considering weather. This suggests that severe weather events, such as storms or flooding causing road closures, have a substantial impact on traffic. Secondly, the crime incident plays an essential role in maintaining performance on METR-LA. This may be due to the high frequency of crimes in downtown LA, which can greatly influence average traffic speed through road closures.

\subsubsection{Fusion Mechanism Design}
We tested the performance using different fusion and alignment mechanisms, as described in Section \ref{sq:fu}. 
"Fuse-Traffic-Gating" means replacing the cross attention block with gating block. "Fuse-Traffic-Add" means directly add the two embeddings without alignment. "Fuse-Traffic-Concatenation" means directly concatenate the two embeddings without alignment.
As shown in Table \ref{tab:ablation}, results demonstrate that concatenation performs poorest, highlighting adding can effectively blending shared information while preserving their interconnections in the latent space. In summary, ablation experiments on three datasets demonstrate that the fusion and alignment mechanism in our model is effective.

\begin{table}[h]
\caption{Ablation Study on Fusion and Alignment Mechanism}
\label{tab:ablation}
\centering{
\setlength{\tabcolsep}{4pt}
\begin{tabular}{lccccc}
\specialrule{1.5pt}{0pt}{0pt} % 双线
Dataset & Model & MAE & RMSE & MAPE\\\hline
\multirow{4}{*}{METR-LA} & Fuse-Traffic & \textbf{2.94} & \textbf{6.03} & \textbf{8.13} \\
&Fuse-Traffic-Gating & 3.04 & 6.19 & 9.67\\
&Fuse-Traffic-Add &3.02 &6.17 &9.16\\
&Fuse-Traffic-Concatenation & 3.18 & 6.54  & 9.59\\
\midrule
\multirow{4}{*}{PEMS-BAY} & Fuse-Traffic & \textbf{1.58} & \textbf{3.55} &\textbf{4.37} \\
&Fuse-Traffic-Gating & 1.65& 3.64&4.80 \\
&Fuse-Traffic-Add & 1.63 & 3.63 &4.74 \\
&Fuse-Traffic-Concatenation & 1.70 & 3.71  & 5.01\\
\midrule
\multirow{4}{*}{PEMS03} & Fuse-Traffic & \textbf{15.05} & \textbf{26.57} &\textbf{15.28} \\
&Fuse-Traffic-Gating & 15.15& 26.84&15.40 \\
&Fuse-Traffic-Add & 15.13 & 26.90 &15.74 \\
&Fuse-Traffic-Concatenation & 15.70 & 27.11  & 16.01\\
\specialrule{1.0pt}{0pt}{0pt}
\end{tabular}
}
\end{table}

% \section{Encoder Attention Analysis(RQ4)}
% We study the attention map according to different event type.

\section{Case Study and Visualization (RQ5)}
\begin{figure}
    \centering
    \includegraphics[width=1\linewidth]{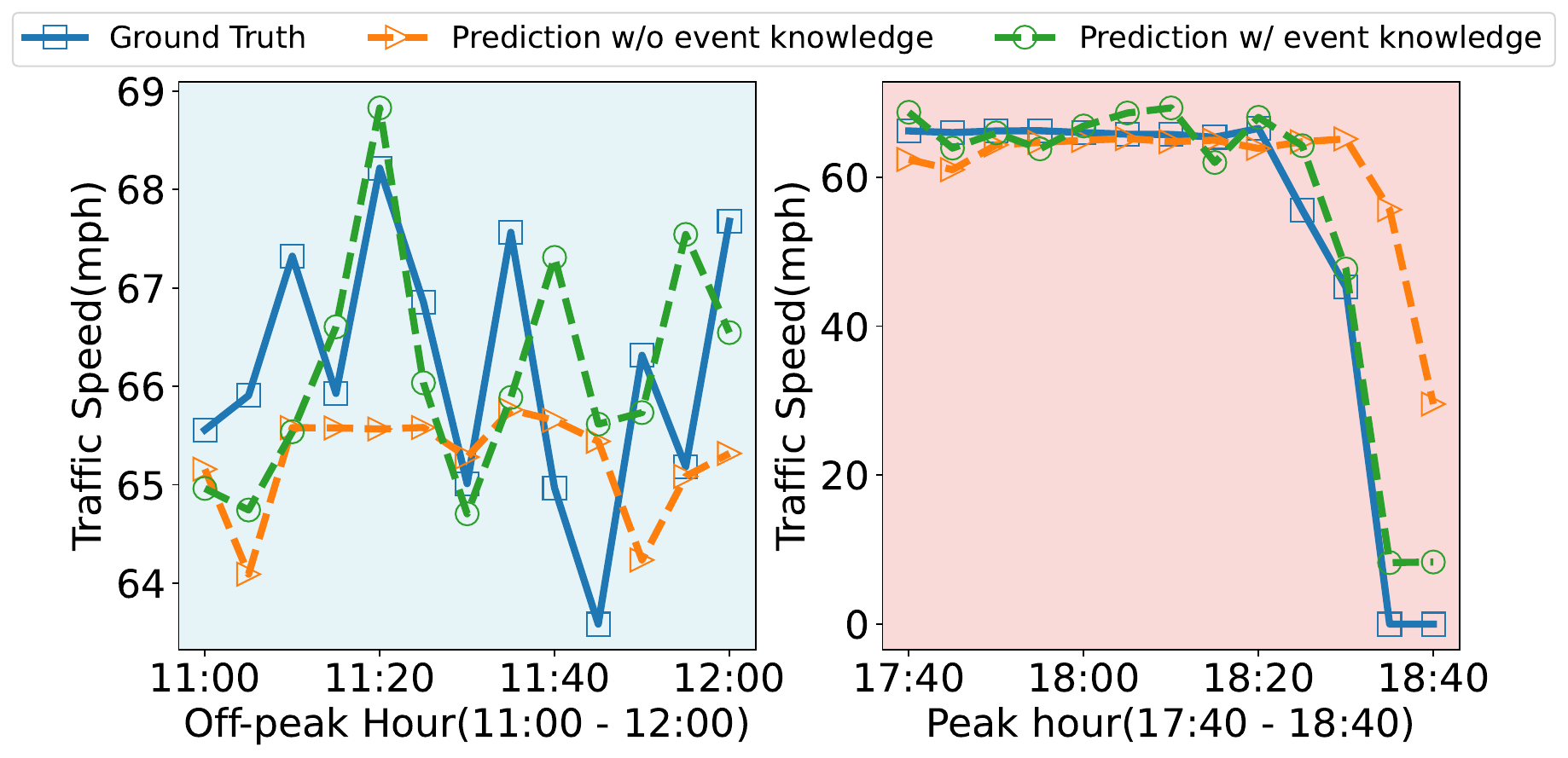}
    \caption{Comparison of Predictions (With/Without Event Knowledge) Against Ground Truth}
    \label{fig:vis_top} % Changed label to be unique
    \vfill % This command adds flexible vertical space, pushing the figures apart.
    \includegraphics[width=1\linewidth]{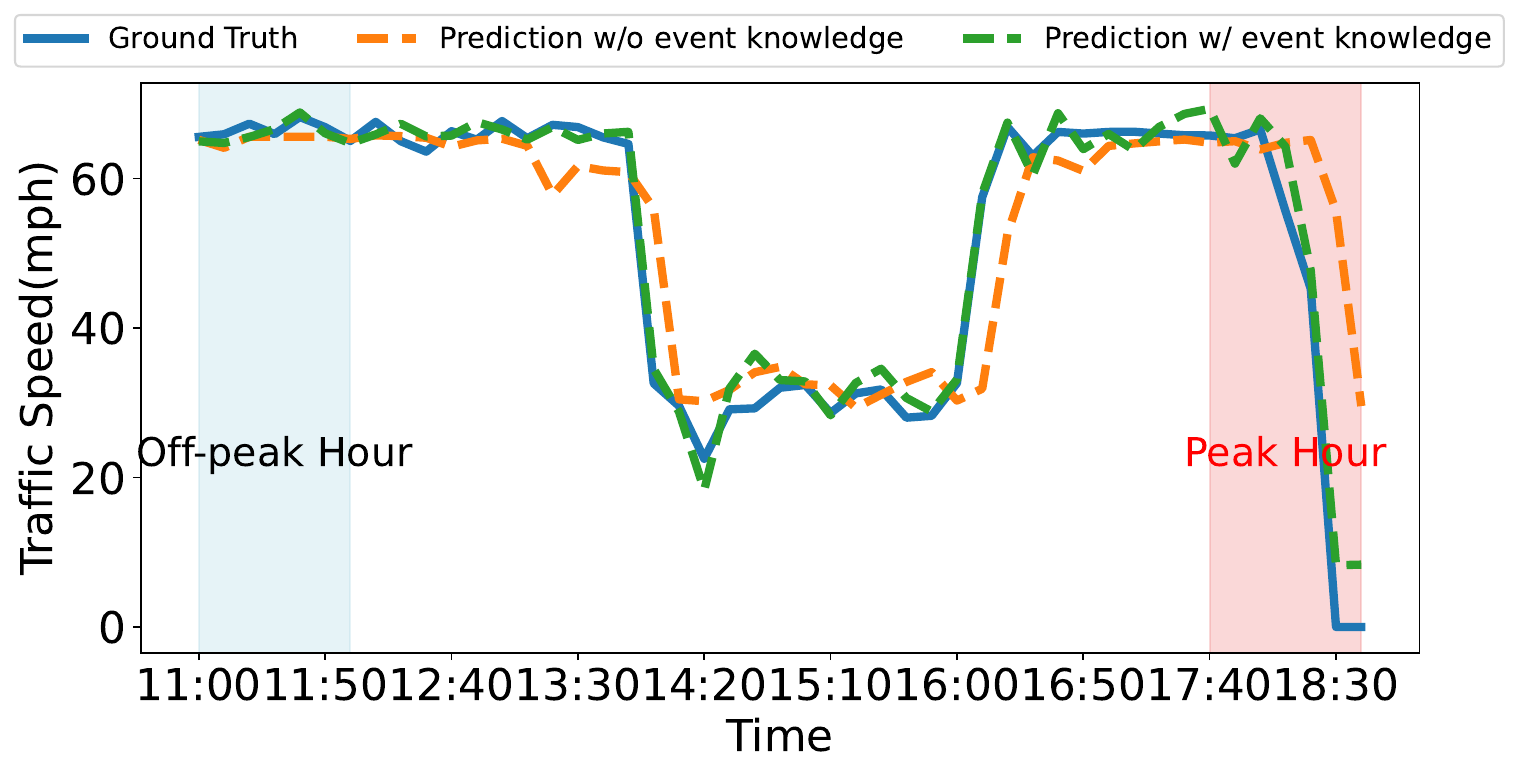}
    \caption{The Traffic Speed Time Series of Sensor 82 at Peak Hour and Off-peak hour}
    \label{fig:vis_bottom}
\end{figure}

To demonstrate the impact of incorporating event knowledge, we present a case study analyzing the traffic speed predictions for sensor 82 from 11:00 to 18:40. The baseline model used is D2STGNN. As shown in Figure \ref{fig:vis_bottom}, the traffic speed for this sensor reveals periods of sharp declines during the evening peak FUSE-Traffic, which was caused by a traffic accident near Sherman Oaks Castle Park.

Figure \ref{fig:vis_top} provides a granular comparison between the predictions of our event-aware model and the baseline model that operates without event knowledge.
shows that during the off-peak FUSE-Traffic (11:00 - 12:00), the ground truth traffic speed exhibits minor fluctuations, varying between approximately 63 mph and 68 mph. The model without event knowledge fails to capture this dynamic, erroneously predicting a nearly constant speed of 66 mph. In contrast, the prediction with event knowledge successfully mirrors the volatility of the ground truth, accurately tracking the dips and rises in speed. During the evening peak hour (17:40 - 18:40), the traffic speed value plummeted from over 60 mph to near zero. The baseline model's prediction shows a decline, but with a significant lag, failing to capture the rapid speed drop seen in the ground truth. In contrast, our event-aware model accurately captures the sharp, downward trend, demonstrating its capability to foresee and adapt to abrupt changes caused by external events. This analysis clearly illustrates that integrating event knowledge is crucial for capturing the complex dynamics of real-world traffic and improving prediction accuracy during anomalous conditions.

\begin{figure}[htbp]
    % \centering
    \includegraphics[width=\linewidth]{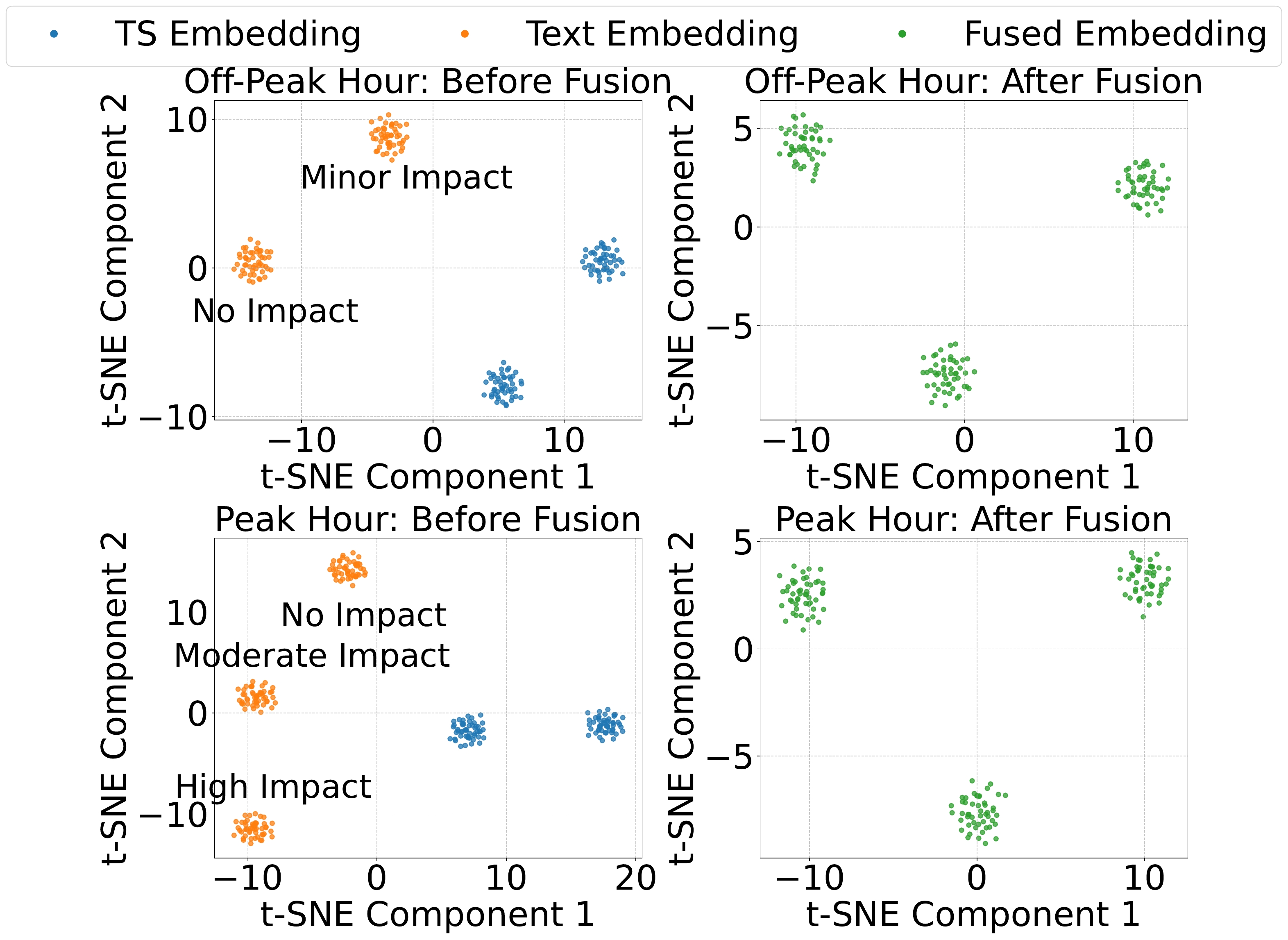}
    \caption{T-SNE Embedding Distributions of Time Series(TS) and Text Modalities}
    \label{fig:tSNE}
\end{figure}

Additionally, we visualize the embeddings of the time series (TS) and text modalities, both before and after fusion in Figure \ref{fig:tSNE}. By reducing the dimensionality of the embedding weights of the trained Fuse-Traffic model from 512 to 2 using t-SNE \cite{van2008visualizing}, we observe the following insights: (1)Without fusion, the TS embedding only forms two distinct clusters and fails to capture the minor event impact that occurred between 11:20 and 11:40. This highlights the limitation of the baseline model in capturing subtle traffic variations during normal traffic conditions. (2) During the peak FUSE-Traffic, the traffic accident cause moderate impact around 18:20, and follows a high impact around 18:30. The TS embedding without fusion forms only two clusters, failing to capture the shift in event impact, which explains why the model does not predict a sharp decline in traffic speed.

\section{Related Work}

\subsection{GNN-based Traffic Forecasting}
Traffic forecasting is an important type of time series forecasting task. Early studies in this domain primarily concentrated on the temporal dynamics of traffic flow, often employing statistical models or shallow machine learning techniques\cite{pan2012utilizing,stock2001vector,moreira2013predicting}. The current
trends of this field revolves around designing cutting-edge spatiotemporal graph neural networks\cite{10242138,10.1145/3627673.3679554,bai2020adaptive,HU2024323,Wang_Adiga_Chen_Sadilek_Venkatramanan_Marathe_2022}. STGCN\cite{Yu2018STGCN} demonstrated the power of combining graph convolutions with temporal convolutional layers to effectively model these intertwined spatial and temporal dependencies. GraphWaveNet\cite{Wu2019GraphWaveNet} introduced the use of adaptive adjacency matrices and dilated causal convolutions to further enhance the ability to capture complex dependencies without relying on predefined graph structures. Subsequently, a plethora of methods have focused on designing more sophisticated spatio-temporal feature extraction modules to achieve superior predictive performance\cite{Fang2023STWave, shao2022decoupled}.

\subsection{GNN-based Traffic Forecasting with Manually Engineered Event Features}
GNN-based models have proven to be effective in capturing regular, periodic patterns in human mobility. However, their performance often degrades when faced with irregular situations characterized by low periodicity, such as those induced by public events, festivals, or unforeseen incidents. Previous research has tended to model the features associated with such anomalous conditions in a relatively coarse-grained manner. For example, STCL\cite{liu2022spatial} utilizes one-hot encoding to indicate accident occurrence. DAMGNet\cite{wu2021dual} uses entity embedding
to encode accident statuses for predicting time delay. Xie et al.\cite{Xie2020DIGCNet} proposed a Deep Incident-Aware Graph Convolutional Network to incorporate traffic incident information along with spatiotemporal data for traffic speed prediction. However, this approach requires manual definition of an Incident Effect Score and manual adjustment of incident severity parameters.  Luo et al.\cite{Luo2023M3AN} constructs distinct subgraphs for different traffic conditions to model node features and address abrupt traffic patterns. Such a strategy, however, relies heavily on domain knowledge for subgraph construction and may not generalize well to different urban layouts or novel event types. A common limitation of the aforementioned models is their dependence on low-dimensional, manually engineered features to represent complex events. This reliance on handcrafted inputs can restrict model generalizability and lead to a significant loss of the nuanced information contained within textual descriptions of public events—a gap the current work needs to address.

\subsection{GNN-based Traffic Forecasting with Event Semantics from Textual Data}
% Multimodal time series analysis seeks to enhance forecasting accuracy by modeling time series data in conjunction with complementary data modalities, such as images, videos, and text. The recent remarkable advancements in language models have catalyzed the development of numerous models that fuse textual data with time series forecasting \cite{jin2024timellm, zhou2023gpt4ts}. Despite these advancements, their application in the domain of traffic forecasting, particularly for predicting traffic patterns influenced by public events, remains relatively limited. 
To better capture the influence of public events, Some prior works have begun to explore the integration of event-related textual data into traffic prediction models. For instance, STORM\cite{Hong2024STORM} incorporates semantic context, which includes the location and impact factors of events such as typhoons, concerts, and sports, along with spatio-temporal data to predict abnormal crowd traffic. T3\cite{EventTraffic2024} addresses event traffic forecasting with sparse multimodal data by employing pre-trained text encoders to generate semantic embeddings from textual descriptions of events. However, a common limitation of such models is their reliance on large-scale, specifically curated event datasets for effective training. 

\subsection{Traffic Forecasting Forecasting with Event
Semantics from LLM}
In contrast, Large Language Models (LLMs), with their extensive world knowledge, can potentially provide crucial context that influences spatio-temporal traffic dynamics without the need for event datasets. UrbanGPT \cite{Li2024UrbanGPT} developes a spatio-temporal LLM framework employing specific encoders and instruction tuning methods to enhance LLM comprehension of spatio-temporal dependencies, aiming for strong generalization across urban tasks, particularly under data scarcity. STD-PLM \cite{huang2025stdplmunderstandingspatialtemporal} adapts pre-trained language models by designing explicit spatial and temporal tokenizers without prompting, enabling effective traffic forecasting and imputation. Despite these advancements in leveraging LLMs for urban spatio-temporal contexts, a critical gap persists: these methods still do not explicitly integrate the nuanced understanding of \textbf{public event data}.  This potential motivates our approach: to leverage the inherent natural language understanding and world knowledge capabilities of LLMs to directly query for, interpret, and integrate relevant event information based on spatial and temporal context, thereby enhancing traffic forecasting models.

% STORM  explicitly integrated semantic context, including event category embeddings, for predicting abnormal crowd traffic. Zhao and Ma \cite{Zhao2022NaiveBayes} used Naïve Bayes, likely informed by schedule data, to classify scenarios (normal vs. planned event) and switch between tailored prediction models for metro flow. AIMEN \cite{AIMEN2024} incorporates the Accident Impact Learning Expert (AILE) to encode detailed accident
% profiles and generate Accident Propagation Graphs that model how accidents influence traffic patterns across road networks.

% These methods often rely on manually engineered low-dimensional features, potentially losing nuanced information present in detailed descriptions. 

% Recently, T3 tackled event traffic forecasting with sparse multimodal data by using pre-trained text encoders to generate semantic embeddings from textual event descriptions. This highlights the potential of leveraging richer text data. However, the vast number and diversity of events make manual collection and processing burdensome. This motivates our approach: to leverage the natural language understanding and world knowledge capabilities of modern LLMs to directly query for, interpret, and integrate relevant event information based on spatial and temporal context. By processing textual descriptions directly, we aim to bypass extensive manual preprocessing and capture richer, more nuanced event semantics to improve spatial-temporal forecasting accuracy.

\section{Conclusion}

In this paper, we proposed FUSE-Traffic, a novel multimodal traffic forecasting framework that integrates structured spatio-temporal traffic data with unstructured event information extracted via Large Language Models (LLMs). By dynamically querying and interpreting contextual event semantics, and fusing them with traffic representations through a fusion and alignment mechanism, FUSE-Traffic effectively addresses the challenge of modeling non-periodic, event-driven traffic dynamics—an aspect where traditional GNN-based methods often underperform.

Extensive experiments on multiple real-world datasets demonstrate that FUSE-Traffic consistently outperforms existing state-of-the-art models across a variety of prediction horizons and event impact levels. Our ablation studies further confirm the importance of both the event-aware prompting strategy and the cross-modality alignment design. The visualization and t-SNE analysis provide intuitive insights into how event knowledge reshapes the feature space and contributes to improved predictive robustness.

%This work opens avenues for deeper integration between foundation models and structured forecasting pipelines in intelligent transportation systems. Future directions include scaling the framework to real-time deployment scenarios, extending to multimodal event sources (e.g., images, social media), and exploring generalization across cities and event domains.

\begin{comment}
\section{Acknowledgments}

Identification of funding sources and other support, and thanks to
individuals and groups that assisted in the research and the
preparation of the work should be included in an acknowledgment
section, which is placed just before the reference section in your
document.
\end{comment}
\newpage
\bibliographystyle{ACM-Reference-Format}
\bibliography{reference}

%%% -*-BibTeX-*-
%%% Do NOT edit. File created by BibTeX with style
%%% ACM-Reference-Format-Journals [18-Jan-2012].

\begin{thebibliography}{23}

%%% ====================================================================
%%% NOTE TO THE USER: you can override these defaults by providing
%%% customized versions of any of these macros before the \bibliography
%%% command.  Each of them MUST provide its own final punctuation,
%%% except for \shownote{} and \showURL{}.  The latter two
%%% do not use final punctuation, in order to avoid confusing it with
%%% the Web address.
%%%
%%% To suppress output of a particular field, define its macro to expand
%%% to an empty string, or better, \unskip, like this:
%%%
%%% \newcommand{\showURL}[1]{\unskip}   % LaTeX syntax
%%%
%%% \def \showURL #1{\unskip}           % plain TeX syntax
%%%
%%% ====================================================================

\ifx \showCODEN    \undefined \def \showCODEN     #1{\unskip}     \fi
\ifx \showISBNx    \undefined \def \showISBNx     #1{\unskip}     \fi
\ifx \showISBNxiii \undefined \def \showISBNxiii  #1{\unskip}     \fi
\ifx \showISSN     \undefined \def \showISSN      #1{\unskip}     \fi
\ifx \showLCCN     \undefined \def \showLCCN      #1{\unskip}     \fi
\ifx \shownote     \undefined \def \shownote      #1{#1}          \fi
\ifx \showarticletitle \undefined \def \showarticletitle #1{#1}   \fi
\ifx \showURL      \undefined \def \showURL       {\relax}        \fi
% The following commands are used for tagged output and should be
% invisible to TeX
\providecommand\bibfield[2]{#2}
\providecommand\bibinfo[2]{#2}
\providecommand\natexlab[1]{#1}
\providecommand\showeprint[2][]{arXiv:#2}

\bibitem[Bai et~al\mbox{.}(2020)]%
        {bai2020adaptive}
\bibfield{author}{\bibinfo{person}{Lei Bai}, \bibinfo{person}{Lina Yao}, \bibinfo{person}{Can Li}, \bibinfo{person}{Xianzhi Wang}, {and} \bibinfo{person}{Can Wang}.} \bibinfo{year}{2020}\natexlab{}.
\newblock \showarticletitle{Adaptive graph convolutional recurrent network for traffic forecasting}.
\newblock \bibinfo{journal}{\emph{Advances in neural information processing systems}}  \bibinfo{volume}{33} (\bibinfo{year}{2020}), \bibinfo{pages}{17804--17815}.
\newblock


\bibitem[Fang et~al\mbox{.}(2023)]%
        {Fang2023STWave}
\bibfield{author}{\bibinfo{person}{Yuchen Fang}, \bibinfo{person}{Yanjun Qin}, \bibinfo{person}{Haiyong Luo}, \bibinfo{person}{Fang Zhao}, \bibinfo{person}{Bingbing Xu}, \bibinfo{person}{Liang Zeng}, {and} \bibinfo{person}{Chenxing Wang}.} \bibinfo{year}{2023}\natexlab{}.
\newblock \showarticletitle{When Spatio-Temporal Meet Wavelets: Disentangled Traffic Forecasting via Efficient Spectral Graph Attention Networks}. In \bibinfo{booktitle}{\emph{2023 IEEE 39th International Conference on Data Engineering (ICDE)}}. \bibinfo{pages}{517--529}.
\newblock


\bibitem[Han et~al\mbox{.}(2024)]%
        {EventTraffic2024}
\bibfield{author}{\bibinfo{person}{Xiao Han}, \bibinfo{person}{Zhenduo Zhang}, \bibinfo{person}{Yiling Wu}, \bibinfo{person}{Xinfeng Zhang}, {and} \bibinfo{person}{Zhe Wu}.} \bibinfo{year}{2024}\natexlab{}.
\newblock \showarticletitle{Event Traffic Forecasting with Sparse Multimodal Data}. In \bibinfo{booktitle}{\emph{Proceedings of the 32nd ACM International Conference on Multimedia}} (Melbourne VIC, Australia) \emph{(\bibinfo{series}{MM '24})}. \bibinfo{publisher}{Association for Computing Machinery}, \bibinfo{address}{New York, NY, USA}, \bibinfo{pages}{8855–8864}.
\newblock
\showISBNx{9798400706868}


\bibitem[Hong et~al\mbox{.}(2024)]%
        {Hong2024STORM}
\bibfield{author}{\bibinfo{person}{Yayao Hong}, \bibinfo{person}{Hang Zhu}, \bibinfo{person}{Tieqi Shou}, \bibinfo{person}{Zeyu Wang}, \bibinfo{person}{Liyue Chen}, \bibinfo{person}{Leye Wang}, \bibinfo{person}{Cheng Wang}, {and} \bibinfo{person}{Longbiao Chen}.} \bibinfo{year}{2024}\natexlab{}.
\newblock \showarticletitle{{STORM:} {A} Spatio-Temporal Context-Aware Model for Predicting Event-Triggered Abnormal Crowd Traffic}.
\newblock \bibinfo{journal}{\emph{IEEE Transactions on Intelligent Transportation Systems}} \bibinfo{volume}{25}, \bibinfo{number}{10} (\bibinfo{year}{2024}), \bibinfo{pages}{13051--13066}.
\newblock


\bibitem[Hu et~al\mbox{.}(2024a)]%
        {10.1145/3627673.3679554}
\bibfield{author}{\bibinfo{person}{Junfeng Hu}, \bibinfo{person}{Xu Liu}, \bibinfo{person}{Zhencheng Fan}, \bibinfo{person}{Yifang Yin}, \bibinfo{person}{Shili Xiang}, \bibinfo{person}{Savitha Ramasamy}, {and} \bibinfo{person}{Roger Zimmermann}.} \bibinfo{year}{2024}\natexlab{a}.
\newblock \showarticletitle{Prompt-Based Spatio-Temporal Graph Transfer Learning}. In \bibinfo{booktitle}{\emph{Proceedings of the 33rd ACM International Conference on Information and Knowledge Management}} (Boise, ID, USA) \emph{(\bibinfo{series}{CIKM '24})}. \bibinfo{publisher}{Association for Computing Machinery}, \bibinfo{address}{New York, NY, USA}, \bibinfo{pages}{890–899}.
\newblock
\showISBNx{9798400704369}
\href{https://doi.org/10.1145/3627673.3679554}{doi:\nolinkurl{10.1145/3627673.3679554}}


\bibitem[Hu et~al\mbox{.}(2024b)]%
        {HU2024323}
\bibfield{author}{\bibinfo{person}{Na Hu}, \bibinfo{person}{Dafang Zhang}, \bibinfo{person}{Kun Xie}, \bibinfo{person}{Wei Liang}, \bibinfo{person}{Kuan-Ching Li}, {and} \bibinfo{person}{Albert~Y. Zomaya}.} \bibinfo{year}{2024}\natexlab{b}.
\newblock \showarticletitle{Dynamic multi-scale spatial–temporal graph convolutional network for traffic flow prediction}.
\newblock \bibinfo{journal}{\emph{Future Generation Computer Systems}}  \bibinfo{volume}{158} (\bibinfo{year}{2024}), \bibinfo{pages}{323--332}.
\newblock
\showISSN{0167-739X}
\href{https://doi.org/10.1016/j.future.2024.04.052}{doi:\nolinkurl{10.1016/j.future.2024.04.052}}


\bibitem[Huang et~al\mbox{.}(2025)]%
        {huang2025stdplmunderstandingspatialtemporal}
\bibfield{author}{\bibinfo{person}{YiHeng Huang}, \bibinfo{person}{Xiaowei Mao}, \bibinfo{person}{Shengnan Guo}, \bibinfo{person}{Yubin Chen}, \bibinfo{person}{Junfeng Shen}, \bibinfo{person}{Tiankuo Li}, \bibinfo{person}{Youfang Lin}, {and} \bibinfo{person}{Huaiyu Wan}.} \bibinfo{year}{2025}\natexlab{}.
\newblock \bibinfo{title}{STD-PLM: Understanding Both Spatial and Temporal Properties of Spatial-Temporal Data with PLM}.
\newblock
\showeprint[arxiv]{2407.09096}~[cs.LG]


\bibitem[Li et~al\mbox{.}(2017)]%
        {li2017diffusion}
\bibfield{author}{\bibinfo{person}{Yaguang Li}, \bibinfo{person}{Rose Yu}, \bibinfo{person}{Cyrus Shahabi}, {and} \bibinfo{person}{Yan Liu}.} \bibinfo{year}{2017}\natexlab{}.
\newblock \showarticletitle{Diffusion convolutional recurrent neural network: Data-driven traffic forecasting}.
\newblock \bibinfo{journal}{\emph{arXiv preprint arXiv:1707.01926}} (\bibinfo{year}{2017}).
\newblock


\bibitem[Li et~al\mbox{.}(2024)]%
        {Li2024UrbanGPT}
\bibfield{author}{\bibinfo{person}{Zhonghang Li}, \bibinfo{person}{Lianghao Xia}, \bibinfo{person}{Jiabin Tang}, \bibinfo{person}{Yong Xu}, \bibinfo{person}{Lei Shi}, \bibinfo{person}{Long Xia}, \bibinfo{person}{Dawei Yin}, {and} \bibinfo{person}{Chao Huang}.} \bibinfo{year}{2024}\natexlab{}.
\newblock \showarticletitle{UrbanGPT: Spatio-Temporal Large Language Models}. In \bibinfo{booktitle}{\emph{Proceedings of the 30th ACM SIGKDD Conference on Knowledge Discovery and Data Mining}} (Barcelona, Spain) \emph{(\bibinfo{series}{KDD '24})}. \bibinfo{publisher}{Association for Computing Machinery}, \bibinfo{address}{New York, NY, USA}, \bibinfo{pages}{5351–5362}.
\newblock
\showISBNx{9798400704901}


\bibitem[Liu et~al\mbox{.}(2022)]%
        {liu2022spatial}
\bibfield{author}{\bibinfo{person}{Zichuan Liu}, \bibinfo{person}{Rui Zhang}, \bibinfo{person}{Chen Wang}, \bibinfo{person}{Zhu Xiao}, {and} \bibinfo{person}{Hongbo Jiang}.} \bibinfo{year}{2022}\natexlab{}.
\newblock \showarticletitle{Spatial-temporal conv-sequence learning with accident encoding for traffic flow prediction}.
\newblock \bibinfo{journal}{\emph{IEEE Transactions on Network Science and Engineering}} \bibinfo{volume}{9}, \bibinfo{number}{3} (\bibinfo{year}{2022}), \bibinfo{pages}{1765--1775}.
\newblock


\bibitem[Luo et~al\mbox{.}(2023)]%
        {Luo2023M3AN}
\bibfield{author}{\bibinfo{person}{Dan Luo}, \bibinfo{person}{Dong Zhao}, \bibinfo{person}{Zijian Cao}, \bibinfo{person}{Mingyao Wu}, \bibinfo{person}{Liang Liu}, {and} \bibinfo{person}{Huadong Ma}.} \bibinfo{year}{2023}\natexlab{}.
\newblock \showarticletitle{M3AN: Multitask Multirange Multisubgraph Attention Network for Condition-Aware Traffic Prediction}.
\newblock \bibinfo{journal}{\emph{IEEE Transactions on Intelligent Transportation Systems}} \bibinfo{volume}{24}, \bibinfo{number}{1} (\bibinfo{year}{2023}), \bibinfo{pages}{218--232}.
\newblock


\bibitem[Moreira-Matias et~al\mbox{.}(2013)]%
        {moreira2013predicting}
\bibfield{author}{\bibinfo{person}{Luis Moreira-Matias}, \bibinfo{person}{Joao Gama}, \bibinfo{person}{Michel Ferreira}, \bibinfo{person}{Joao Mendes-Moreira}, {and} \bibinfo{person}{Luis Damas}.} \bibinfo{year}{2013}\natexlab{}.
\newblock \showarticletitle{Predicting taxi--passenger demand using streaming data}.
\newblock \bibinfo{journal}{\emph{IEEE Transactions on Intelligent Transportation Systems}} \bibinfo{volume}{14}, \bibinfo{number}{3} (\bibinfo{year}{2013}), \bibinfo{pages}{1393--1402}.
\newblock


\bibitem[Pan et~al\mbox{.}(2012)]%
        {pan2012utilizing}
\bibfield{author}{\bibinfo{person}{Bei Pan}, \bibinfo{person}{Ugur Demiryurek}, {and} \bibinfo{person}{Cyrus Shahabi}.} \bibinfo{year}{2012}\natexlab{}.
\newblock \showarticletitle{Utilizing real-world transportation data for accurate traffic prediction}. In \bibinfo{booktitle}{\emph{2012 ieee 12th international conference on data mining}}. IEEE, \bibinfo{pages}{595--604}.
\newblock


\bibitem[Qian et~al\mbox{.}(2024)]%
        {10242138}
\bibfield{author}{\bibinfo{person}{Weizhu Qian}, \bibinfo{person}{Yan Zhao}, \bibinfo{person}{Dalin Zhang}, \bibinfo{person}{Bowei Chen}, \bibinfo{person}{Kai Zheng}, {and} \bibinfo{person}{Xiaofang Zhou}.} \bibinfo{year}{2024}\natexlab{}.
\newblock \showarticletitle{Towards a Unified Understanding of Uncertainty Quantification in Traffic Flow Forecasting}.
\newblock \bibinfo{journal}{\emph{IEEE Transactions on Knowledge and Data Engineering}} \bibinfo{volume}{36}, \bibinfo{number}{5} (\bibinfo{year}{2024}), \bibinfo{pages}{2239--2256}.
\newblock
\href{https://doi.org/10.1109/TKDE.2023.3312261}{doi:\nolinkurl{10.1109/TKDE.2023.3312261}}


\bibitem[Shao et~al\mbox{.}(2022)]%
        {shao2022decoupled}
\bibfield{author}{\bibinfo{person}{Zezhi Shao}, \bibinfo{person}{Zhao Zhang}, \bibinfo{person}{Wei Wei}, \bibinfo{person}{Fei Wang}, \bibinfo{person}{Yongjun Xu}, \bibinfo{person}{Xin Cao}, {and} \bibinfo{person}{Christian~S Jensen}.} \bibinfo{year}{2022}\natexlab{}.
\newblock \showarticletitle{Decoupled dynamic spatial-temporal graph neural network for traffic forecasting}.
\newblock \bibinfo{journal}{\emph{Proceedings of the VLDB Endowment}} \bibinfo{volume}{15}, \bibinfo{number}{11} (\bibinfo{year}{2022}), \bibinfo{pages}{2733--2746}.
\newblock


\bibitem[Song et~al\mbox{.}(2020)]%
        {Song_Lin_Guo_Wan_2020}
\bibfield{author}{\bibinfo{person}{Chao Song}, \bibinfo{person}{Youfang Lin}, \bibinfo{person}{Shengnan Guo}, {and} \bibinfo{person}{Huaiyu Wan}.} \bibinfo{year}{2020}\natexlab{}.
\newblock \showarticletitle{Spatial-Temporal Synchronous Graph Convolutional Networks: A New Framework for Spatial-Temporal Network Data Forecasting}.
\newblock \bibinfo{journal}{\emph{Proceedings of the AAAI Conference on Artificial Intelligence}} \bibinfo{volume}{34}, \bibinfo{number}{01} (\bibinfo{date}{Apr.} \bibinfo{year}{2020}), \bibinfo{pages}{914--921}.
\newblock


\bibitem[Stock and Watson(2001)]%
        {stock2001vector}
\bibfield{author}{\bibinfo{person}{James~H Stock} {and} \bibinfo{person}{Mark~W Watson}.} \bibinfo{year}{2001}\natexlab{}.
\newblock \showarticletitle{Vector autoregressions}.
\newblock \bibinfo{journal}{\emph{Journal of Economic perspectives}} \bibinfo{volume}{15}, \bibinfo{number}{4} (\bibinfo{year}{2001}), \bibinfo{pages}{101--115}.
\newblock


\bibitem[Van~der Maaten and Hinton(2008)]%
        {van2008visualizing}
\bibfield{author}{\bibinfo{person}{Laurens Van~der Maaten} {and} \bibinfo{person}{Geoffrey Hinton}.} \bibinfo{year}{2008}\natexlab{}.
\newblock \showarticletitle{Visualizing data using t-SNE.}
\newblock \bibinfo{journal}{\emph{Journal of machine learning research}} \bibinfo{volume}{9}, \bibinfo{number}{11} (\bibinfo{year}{2008}).
\newblock


\bibitem[Wang et~al\mbox{.}(2022)]%
        {Wang_Adiga_Chen_Sadilek_Venkatramanan_Marathe_2022}
\bibfield{author}{\bibinfo{person}{Lijing Wang}, \bibinfo{person}{Aniruddha Adiga}, \bibinfo{person}{Jiangzhuo Chen}, \bibinfo{person}{Adam Sadilek}, \bibinfo{person}{Srinivasan Venkatramanan}, {and} \bibinfo{person}{Madhav Marathe}.} \bibinfo{year}{2022}\natexlab{}.
\newblock \showarticletitle{CausalGNN: Causal-Based Graph Neural Networks for Spatio-Temporal Epidemic Forecasting}.
\newblock \bibinfo{journal}{\emph{Proceedings of the AAAI Conference on Artificial Intelligence}} \bibinfo{volume}{36}, \bibinfo{number}{11} (\bibinfo{date}{Jun.} \bibinfo{year}{2022}), \bibinfo{pages}{12191--12199}.
\newblock
\href{https://doi.org/10.1609/aaai.v36i11.21479}{doi:\nolinkurl{10.1609/aaai.v36i11.21479}}


\bibitem[Wu et~al\mbox{.}(2021)]%
        {wu2021dual}
\bibfield{author}{\bibinfo{person}{I-Ying Wu}, \bibinfo{person}{Fandel Lin}, {and} \bibinfo{person}{Hsun-Ping Hsieh}.} \bibinfo{year}{2021}\natexlab{}.
\newblock \showarticletitle{Dual-Attention Multi-Scale Graph Convolutional Networks for Highway Accident Delay Time Prediction}. In \bibinfo{booktitle}{\emph{Proceedings of the 29th International Conference on Advances in Geographic Information Systems}}. \bibinfo{pages}{554--563}.
\newblock


\bibitem[Wu et~al\mbox{.}(2019)]%
        {Wu2019GraphWaveNet}
\bibfield{author}{\bibinfo{person}{Zonghan Wu}, \bibinfo{person}{Shirui Pan}, \bibinfo{person}{Guodong Long}, \bibinfo{person}{Jing Jiang}, {and} \bibinfo{person}{Chengqi Zhang}.} \bibinfo{year}{2019}\natexlab{}.
\newblock \showarticletitle{Graph WaveNet for Deep Spatial-Temporal Graph Modeling}. In \bibinfo{booktitle}{\emph{Proceedings of the Twenty-Eighth International Joint Conference on Artificial Intelligence, {IJCAI} 2019}}. \bibinfo{pages}{3707--3713}.
\newblock


\bibitem[Xie et~al\mbox{.}(2021)]%
        {Xie2020DIGCNet}
\bibfield{author}{\bibinfo{person}{Pengfei Xie}, \bibinfo{person}{Ting Li}, \bibinfo{person}{Xiqun Li}, \bibinfo{person}{Fuliang Wang}, \bibinfo{person}{Siyuan Sun}, \bibinfo{person}{Wang-Chien Lee}, {and} \bibinfo{person}{Ge Yu}.} \bibinfo{year}{2021}\natexlab{}.
\newblock \showarticletitle{Deep Graph Convolutional Networks for Incident-Driven Traffic Speed Prediction}.
\newblock \bibinfo{journal}{\emph{IEEE Transactions on Intelligent Transportation Systems}} \bibinfo{volume}{22}, \bibinfo{number}{6} (\bibinfo{year}{2021}), \bibinfo{pages}{3549--3563}.
\newblock


\bibitem[Yu et~al\mbox{.}(2018)]%
        {Yu2018STGCN}
\bibfield{author}{\bibinfo{person}{Bing Yu}, \bibinfo{person}{Haoteng Yin}, {and} \bibinfo{person}{Zhanxing Zhu}.} \bibinfo{year}{2018}\natexlab{}.
\newblock \showarticletitle{Spatio-Temporal Graph Convolutional Networks: {A} Deep Learning Framework for Traffic Forecasting}. In \bibinfo{booktitle}{\emph{Proceedings of the Twenty-Seventh International Joint Conference on Artificial Intelligence, {IJCAI} 2018}}. \bibinfo{pages}{3634--3640}.
\newblock


\end{thebibliography}

\newpage
\appendix

% \section{Math Notations}

% \begin{table}[ht]
% \caption{Summary of Notations}
% \label{tab:notations}
% \small
% \begin{tabularx}{\columnwidth}{@{} l X @{}}
% \toprule
% \textbf{Symbol / Term} & \textbf{Description} \\
% \midrule
% $\mathcal{G} = (\mathcal{V}, \mathcal{A})$ & Road network graph; $\mathcal{V}$ is the set of sensors, $\mathcal{A} \in \mathbb{R}^{N \times N}$ is the adjacency matrix. \\
% $N$ & Number of sensors (sensors) in the road network. \\
% $T$ & Total number of discrete time intervals. \\
% $X \in \mathbb{R}^{N \times T}$ & Traffic status matrix, where $X_t = (x_t^{(1)}, x_t^{(2)}, \dots, x_t^{(N)})$. \\
% $x_t^{(i)}$ & Traffic status (e.g., speed or flow) at node $i$ and time $t$. \\
% $H_{\text{in}}$ & Number of past time steps used as input. \\
% $H_{\text{out}}$ & Number of future time steps to predict. \\
% $Y_{t+1}, \dots, Y_{t+H_{\text{out}}}$ & Ground-truth traffic status in prediction window. \\
% $f(\cdot)$ & Prediction function mapping history to future traffic. \\
% $d$ & Dimension of embeddings. \\
% $E_s \in \mathbb{R}^{N \times d}$ & Spatial embedding of sensors. \\
% $E_{\text{tod}} \in \mathbb{R}^{T_1 \times d}$ & Temporal embedding of time-of-day. \\
% $E_{\text{dow}} \in \mathbb{R}^{T_2 \times d}$ & Temporal embedding of day-of-week. \\
% $E$ & Concatenated embedding: $E = \text{FC}(X) \, \| \, E_s \, \| \, E_{\text{tod}} \, \| \, E_{\text{dow}}$. \\
% $\text{FC}(\cdot)$ & Fully connected (dense) layer. \\
% \bottomrule
% \end{tabularx}
% \end{table}

\end{document}